\documentclass{article}

    \PassOptionsToPackage{numbers, compress}{natbib}

 \usepackage[main, final]{neurips_2026}

\usepackage[utf8]{inputenc} 
\usepackage[T1]{fontenc}    
\usepackage{hyperref}       
\usepackage{url}            
\usepackage{booktabs}       
\usepackage{amsfonts}       
\usepackage{nicefrac}       
\usepackage{microtype}      
\usepackage{xcolor}         
\usepackage{graphicx}       
\usepackage{subcaption}     
\usepackage{multirow}       
\usepackage{colortbl}       
\usepackage{amsmath}        
\usepackage{amssymb}        
\usepackage{mathtools}      
\usepackage{wrapfig}

\usepackage{algorithm}
\usepackage{algorithmic}
\renewcommand{\algorithmiccomment}[1]{\quad$\triangleright$ #1}
\definecolor{highlightblue}{HTML}{E6F0FF}
\newcommand{\hlc}[1]{\cellcolor{highlightblue}#1}

\title{Latent Action Reparameterization for Efficient Agent Inference}

\author{%
  \textbf{Wenhao Huang}$^{1,8,*}$,
  \textbf{Qingwen Zeng}$^{2,10,*}$,
  \textbf{Qiyue Chen}$^{2}$,
  \textbf{Zijie Guo}$^{3}$,
  \textbf{Yu Sun}$^{9}$,
  \textbf{Cheng Yang}$^{10}$, \\
  \textbf{Siru Ouyang}$^{4}$,
  \textbf{Jiri Gesi}$^{5}$,
  \textbf{Fang Wu}$^{6}$,
  \textbf{Jiayi Zhang}$^{7,10}$,
  \textbf{Huaming Chen}$^{2}$,
  \textbf{Bang Liu}$^{1,8}$, \\
  \textbf{Xiangru Tang}$^{9,\dagger}$,
  \textbf{Chenglin Wu}$^{10,\dagger}$
  \\[-0.1em]
   \small
  $^{1}$Universit\'{e} de Montr\'{e}al \enspace
  $^{2}$The University of Sydney \enspace
  $^{3}$Fudan University \\
  $^{4}$University of Illinois Urbana-Champaign \enspace
  $^{5}$Amazon Science \enspace
  $^{6}$Stanford University \\
  $^{7}$The Hong Kong University of Science and Technology (Guangzhou) \\
  $^{8}$Mila - Quebec AI Institute \enspace
  $^{9}$Yale University \enspace
  $^{10}$DeepWisdom
  }

\begin{document}

\maketitle

\begingroup
\renewcommand{\thefootnote}{*}
\footnotetext{Equal contribution. Author ordering is random.}
\endgroup

\begingroup
\renewcommand{\thefootnote}{$\dagger$}
\footnotetext{Correspondence to Xiangru Tang \texttt{<xiangru.tang@yale.edu>} and Chenglin Wu \texttt{<alexanderwu@deepwisdom.ai>}.}
\endgroup
\setcounter{footnote}{0}

\begin{abstract}
  Large language model (LLM) agents often rely on long sequences of low-level textual actions, resulting in large effective decision horizons and high inference cost. While prior work has focused on improving inference efficiency through system-level optimizations or prompt engineering, we argue that a key bottleneck lies in the representation of the action space itself. We propose Latent Action Reparameterization (LAR), a framework that learns a compact latent action space in which each latent action corresponds to a multi-step semantic behavior. By reparameterizing agent actions into latent units, LAR enables decision making over a shorter effective horizon while preserving the expressiveness of the original action space. Unlike hand-crafted macros or hierarchical controllers, latent actions are learned from agent trajectories and integrated directly into the model, allowing both planning and execution to operate over abstract action representations. Across a range of LLM-based agent benchmarks, LAR significantly reduces the effective action horizon and improves inference efficiency under fixed compute budgets. As a consequence, our approach achieves substantial reductions in action tokens and corresponding wall-clock inference time, while maintaining or improving task success rates. These results suggest that action representation learning is a critical and underexplored factor in scaling efficient LLM agent inference, complementary to advances in model architecture and hardware. Open code is available here~\footnote{\url{https://github.com/EZ-hwh/LAR}}.
\end{abstract}

\vspace{-0.5em}
\section{Introduction}

Large language model (LLM) agents have emerged as a powerful paradigm for solving tasks involving multi-step reasoning, tool use, and interaction with external environments~\citep{wang2024survey}. By repeatedly generating actions conditioned on intermediate observations, LLM agents can perform search, planning, and decision making across diverse domains~\citep{liu2023agentbench,gioacchini2024agentquest, yang2025multi}. However, as these agents are applied to increasingly complex tasks, inference efficiency has become a critical bottleneck~\citep{liu2025costbench}. Agent execution often requires long sequences of decisions, leading to inference latency and prohibitive computational cost, which in turn limits scalability, deployment, and real-time interaction~\citep{gonzalez2025robotouille}.

Prior work has primarily addressed agent inference efficiency through improvements in model architecture, hardware acceleration, system-level optimizations, or prompt engineering~\citep{cai2025fastmtp,chen2024hardware,debnath2025comprehensive,wan2023efficient}. These approaches reduce the cost of individual inference steps or improve throughput, but they largely operate orthogonally to the structure of the agent's decision process itself~\citep{zhou2024survey}. In particular, while per-token generation may become faster~\citep{gim2024prompt,cai2025fastmtp}, the number of decision steps required to complete a task often remains unchanged. As a result, the overall inference cost continues to scale poorly with task horizon, especially in settings that require multi-step reasoning or search~\citep{chen2025towards}.

In this work, we argue that inference efficiency in LLM agents is fundamentally constrained by the representation of the action space, particularly in sequential decision-making settings. In current agent systems, actions are typically realized as low-level textual outputs, where each generated token constitutes an explicit decision that conditions subsequent computation, planning, or interaction with the environment~\citep{kim2025reflact}. Such token-level action representations induce excessively fine-grained decision making, resulting in unnecessarily large effective decision horizons even for semantically simple behaviors~\citep{zhai2025enhancing,chen2024efficient}. Consequently, inference scaling is dominated not by model size alone, but by the granularity at which agent actions are represented and composed over time~\citep{zheng2024prise,yang2025aria}. We therefore posit that action representation should be treated as a first-class modeling choice in LLM-based agents, on par with model architecture and system-level design.

Motivated by this observation, we propose Latent Action Reparameterization (LAR), a framework that learns a compact latent action space for LLM agents. In LAR, each latent action corresponds to a multi-step semantic behavior that would otherwise be realized through a sequence of low-level textual actions. By reparameterizing agent decisions into these latent units, planning and execution can operate over a substantially shorter effective horizon while preserving the expressiveness of the original action space. Unlike hand-crafted macros or hierarchical controllers~\citep{al2015hierarchical,amato2019modeling,bacon2017option}, latent actions in LAR are learned directly from agent trajectories and integrated into the model, enabling end-to-end decision making over abstract yet executable action representations.

A challenge in action abstraction for LLM agents lies in balancing representational abstraction with action executability~\citep{yao2022react,schick2023toolformer}. Fully implicit latent representations are effective for internal reasoning and credit assignment, but they are insufficient for agent systems that must interact with external tools or environments through explicit, protocol-constrained interfaces~\citep{schick2023toolformer,hafner2023mastering}. In such settings, actions must remain decodable, interpretable, and executable by downstream systems. LAR addresses this challenge by explicitly modeling the latent--explicit boundary: latent actions provide higher-level abstraction while remaining directly realizable as concrete, executable action sequences. Specifically, our framework compresses low-entropy, structurally recurring patterns, including system prompts, tool invocation syntax, and recurring configurations into latent units, while strictly preserving high-entropy, parameter-rich inputs (e.g., specific search queries or numerical entities) in the explicit output space. This design reflects a broader principle in agent modeling: increased abstraction is not always beneficial, as executability fundamentally constrains useful action representations~\citep{yao2022react}.

Our contributions are fourfold. \textbf{First}, Significant Efficiency Gains: We demonstrate that LAR significantly reduces the effective action horizon, leading to substantial reductions in action tokens and corresponding system-level gains in token throughput and peak GPU memory across diverse LLM agent benchmarks (Section~\ref{sec:main_results}, Table~\ref{tab:efficiency}). \textbf{Second}, Preserved Task Performance: Despite operating over a compressed latent action space, our approach maintains or improves task success rates compared to baselines using raw textual actions (Table~\ref{tab:main_results}) and transfers to held-out benchmarks without retraining (Section~\ref{sec:cross_benchmark}), proving that efficiency need not come at the cost of performance. \textbf{Third}, Analysis of Abstraction Limits: We identify a distinct performance collapse threshold, empirically delineating the boundary between compressible structural redundancy and essential semantic content (Section~\ref{sec:progressive}). \textbf{Fourth}, New Perspective on Scaling: Our results highlight action representation learning as a critical and underexplored factor in scaling efficient LLM agent inference, validated across model scales up to 32B (Appendix~\ref{app:scalability}) and industrial agent runtimes (Appendix~\ref{app:Openclaw}), offering a complementary path to advances in model architecture and hardware.

\vspace{-0.5em}
\section{Related Work}

\vspace{-0.5em}
A large body of prior work improves the efficiency of LLM-based agents by modifying stages of the agent pipeline, including how inputs are conditioned, how tokens are generated, and how interaction histories are maintained. \textbf{Prompting and Input-Level Control:} Prompting and input-level methods improve efficiency by shaping the conditioning signal before or during inference~\citep{debnath2025comprehensive}. Techniques such as Chain-of-Thought elicit intermediate reasoning that improves accuracy but often increases generation length and latency~\citep{wei2022chain,wang2022self}. Subsequent prompt engineering constrains reasoning formats to reduce verbosity while preserving answer quality~\citep{zhou2022least,li2023guiding, zengbootstrapped}. \textbf{Token-Level Generation Control and Inference-Time Interventions:} Another line of work regulates token emission during inference to reduce redundant generation~\citep{leviathan2023fast,kim2023critic,shridhar2023distilling}. Representative approaches include in-generation guidance that encourages shorter reasoning traces (e.g., ConciseHint-style methods~\citep{tang2025concisehint}) and token scoring or pruning mechanisms that skip low-utility tokens (e.g., TokenSkip~\citep{xia2025tokenskip}). These methods optimize efficiency within the original token-level generation process. \textbf{Context and Memory Optimization for Agents:} For interactive and tool-using agents, efficiency bottlenecks often stem from long interaction histories carried as context~\citep{shinn2023reflexion,park2023generative,packer2023memgpt,zhang2025survey, yu2026dual}. Context and memory optimization methods reduce conditioning costs by compressing or summarizing histories. ACON is the representative optimizing the agent's memory representation via history and observation compression~\citep{kang2025acon}.

Collectively, the above approaches improve efficiency while leaving the decision interface fundamentally unchanged: the agent still reasons and acts at the level of token emissions, and efficiency gains arise from modifying inputs, regulating token generation, or compressing memory~\citep{wan2023efficient}. As a result, the effective decision horizon remains dictated by token-level granularity. In contrast, our work challenges this assumption and targets inefficiency at its source by redefining the unit of decision-making itself. LAR reparameterizes the action space by collapsing multi-step action segments that induce transition-equivalent behaviors into single latent actions, thereby directly reducing the effective decision horizon. Crucially, reparameterization is constrained by executability: parameter-binding actions that determine environment-facing semantics are preserved explicitly, while only stable, context-invariant scaffolds are abstracted. This reframes efficiency not as a byproduct of shorter text, but as a consequence of operating over a more appropriate decision representation.

\begin{figure*}[!t]
\centering
\includegraphics[width=\linewidth]{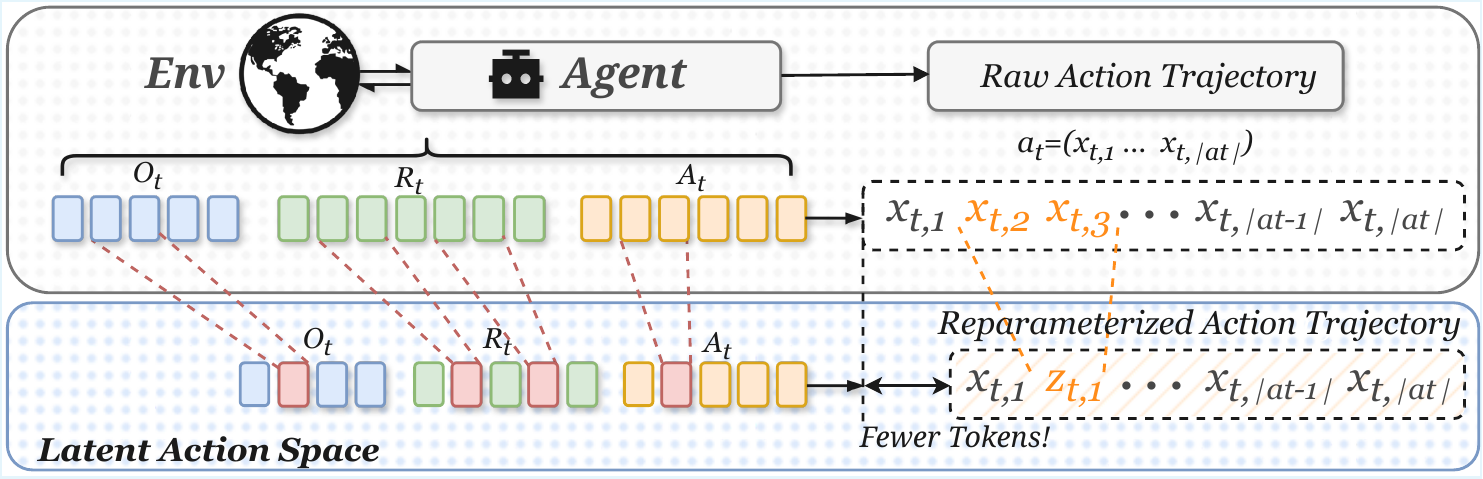}
\caption{
Overview of Latent Action Reparameterization (LAR).
LAR reformulates agent decision making by collapsing transition-equivalent action segments into executable latent actions, thereby reducing the effective decision horizon.
Low-entropy structural components are abstracted into latent actions, while high-entropy, parameter-binding content remains explicit to preserve executability.
}

\vspace{-1.3em}
\label{fig:latent_overview}
\end{figure*}

\vspace{-0.7em}
\section{Methodology}
\label{sec:method}

\vspace{-0.5em}
\subsection{Action Definition and Problem Setup}
\label{sec:problem_setup}

\vspace{-0.3em}
We consider an LLM-based agent in a sequential decision-making setting. An agent--environment interaction is represented as a trajectory $\tau = (o_1, a_1, o_2, a_2, \dots, o_T, a_T)$, where $o_t$ denotes the observation at step $t$ and $a_t$ denotes the action produced by the agent at that step. Observations may include textual context, intermediate reasoning states, tool outputs, or environment feedback.

In contemporary LLM agents, actions are instantiated as explicit textual outputs. Each action $a_t$ is a sequence of generated tokens $a_t = (x_{t,1}, \dots, x_{t,|a_t|})$, with $x_{t,i}$ drawn from the model vocabulary. We treat all generated tokens that condition subsequent computation or interaction as action decisions, including system-level configurations and interaction scaffolds.

We define the effective action horizon of a trajectory as $H_{\mathrm{eff}}(\tau) = \sum_{t=1}^{T} |a_t|$, which measures the number of generation decisions and directly determines inference cost. Our objective is to reduce this horizon by altering action representation, without modifying agent behavior or executability.

\vspace{-0.7em}
\subsection{Latent Action Reparameterization}
\label{sec:reparam}

\vspace{-0.4em}
We propose Latent Action Reparameterization (LAR), which reformulates agent decision making over a compact action space as shown in Fig.~\ref{fig:latent_overview}. Instead of operating over token-level action primitives, LAR enables the agent to reason over higher-level action units that correspond to multi-step semantic behaviors. Let $\mathcal{Z}$ denote the latent action space, and let $z_t \in \mathcal{Z}$ denote the latent action instantiated at step $t$. Each latent action represents a semantic decision unit that subsumes a sequence of low-level actions. Under this representation, the effective horizon becomes $H_{\mathrm{lat}}(\tau) = \sum_t |z_t|$, where each latent action is treated as a single decision step.

LAR aims to preserve the functional behavior induced by original trajectories while eliminating redundant decision steps caused by overly fine-grained action representations, a property we empirically verify in Section~\ref{sec:equivalence}. Planning and execution, therefore operate directly over latent actions rather than token-level primitives.

\vspace{-0.5em}
\subsection{Learning Latent Actions from Trajectories}
\label{sec:learning}

Latent actions are learned directly from agent trajectories by identifying recurrent multi-step behaviors that function as stable decision units. Rather than treating each action $a_t$ as atomic, we consider \emph{action segments}, which may span multiple decision steps or structured sub-sequences within a single action. Such segments capture extended behaviors that recur across trajectories.

We characterize recurrence using \emph{transition equivalence}. Let $\mathcal{T}$ denote the transition dynamics induced by the agent and environment. Two segments $a$ and $a'$ are said to be transition-equivalent if, for any preceding history $h$, the induced transitions satisfy $\mathcal{T}(h \circ a) \approx \mathcal{T}(h \circ a')$, where $\circ$ denotes sequence concatenation and $\approx$ denotes equivalence up to task-relevant outcomes. In practice, this equivalence is not enforced universally but is approximated on the empirical trajectory distribution induced by the agent, with the approximation procedure given in Section~\ref{sec:implementation}.

A latent action $z \in \mathcal{Z}$ is defined as an \emph{equivalence class} of segments under this relation. To integrate latent actions into the agent, each $z$ is parameterized as a vocabulary symbol (Section~\ref{sec:implementation}), so that planning and execution operate over latent actions as over ordinary tokens.

\vspace{-0.5em}
\subsection{Executable Latent Actions}
\label{sec:exec}

Not all latent actions are executable. In our framework, executability subsumes both syntactic validity under external interfaces and semantic correctness with respect to agent--environment interaction. Specifically, while actions must remain decodable and interpretable by downstream systems, true executability further requires that replacing concrete trajectory segments with a latent action does not alter the induced transition behavior.

Formally, a latent action $z$ is executable if all segments belonging to $z$ are transition-equivalent: for any two realizations $a, a' \in z$ and any preceding history $h$, $\mathcal{T}(h \circ a) \approx \mathcal{T}(h \circ a')$. This defines a \emph{semantic constraint} on the latent action space. Latent actions satisfying this constraint correspond to behaviors whose effects are invariant across contexts; in contrast, segments whose effects depend on task-specific parameters or bindings violate transition equivalence and cannot be safely abstracted. The implementation of this constraint, via an entropy-based filter, is described in Section~\ref{sec:implementation}.

\vspace{-0.5em}
\subsection{Implementation}
\label{sec:implementation}

LAR is realized as a four-stage pipeline that operationalizes the formal concepts introduced above: (1) identifying transition-equivalent action segments from agent trajectories, (2) constructing a latent action vocabulary, (3) preparing dual-format training data, and (4) aligning the model's predictive behavior via trajectory-level distillation.

\textbf{Identifying transition-equivalent segments.}
Direct verification of the transition equivalence condition $\mathcal{T}(h \circ a) \approx \mathcal{T}(h \circ a')$ from Section~\ref{sec:learning} over histories $h$ is intractable. We approximate it through the next-token entropy of a candidate segment $s$, defined as $H(s) = -\sum_{w \in V_s} p(w \mid s)\log_2 p(w \mid s)$, where $V_s$ is the set of tokens observed after $s$ and $p(w \mid s)$ is the empirical probability of $w$ following $s$. A low $H(s)$ indicates that the continuation behavior of $s$ is predictable regardless of preceding history, precisely the condition required by transition equivalence. Conversely, high-entropy segments such as specific search queries or task-specific arguments exhibit context-dependent continuations and would violate executability if abstracted. Next-token entropy thus serves as a tractable empirical surrogate for transition equivalence.

The identification procedure (Algorithm~\ref{alg:identify}, Appendix~\ref{app:algorithm}) extracts word-level $n$-grams within boundaries, filters them by frequency ($\text{freq}(s) \geq f_{\min}$) and entropy ($H(s) \leq H_{\max}$), ranks candidates by $\text{score}(s) = \text{freq}(s)/(H(s)+1)$, deduplicates the result, and retains the top-$K$ segments as the latent action set $\mathcal{Z}$. Per-task thresholds and vocabulary sizes are reported in Appendix~\ref{app:lar-config}.

\textbf{Vocabulary and training data.}
Each segment in $\mathcal{Z}$ is assigned a dedicated vocabulary symbol. Training data is prepared in a \emph{dual-trajectory} format: each original trajectory $\tau$ is paired with its reparameterized counterpart $\hat{\tau}$, in which segments matching $\mathcal{Z}$ are replaced by the corresponding latent action symbols via longest-first matching. The original trajectory serves as the teacher signal, while the reparameterized trajectory is the student input.

\textbf{Trajectory-level distillation.}
A frozen copy of the original LLM (the \emph{teacher}) processes $\tau$, while a \emph{student}, the same base model augmented with a LoRA adapter (rank $r=8$, $\alpha=16$, applied to q/k/v/o projections) and new latent action embeddings, processes $\hat{\tau}$. Only the LoRA weights and new embeddings are trainable, amounting to $0.1\%$ of total parameters; all pretrained weights remain frozen. The training objective is pure KL distillation over shared content positions, $\mathcal{L}_{\text{KL}} = \frac{1}{|M|}\sum_{i \in M} D_{\text{KL}}(\mathrm{softmax}(z^T_i/\tau) \,\|\, \mathrm{softmax}(z^S_i/\tau))$, where $M$ is the set of token positions whose textual content is identical in both teacher and student sequences (excluding latent action symbols), $z^T_i$ and $z^S_i$ are the teacher and student logits, and $\tau=2.0$ is the distillation temperature. Restricting the loss to $M$ is the mechanism by which latent action embeddings acquire their semantic content: the student must reproduce the teacher's predictive distribution on non-compressed content despite receiving compressed input, forcing the new embeddings to encode the full semantics of the segments they replace. Detailed training hyperparameters are reported in Appendix~\ref{app:reproducibility}.

\textbf{Inference.}
In inference, latent action symbols are processed identically to ordinary vocabulary tokens through the same embedding lookup and transformer forward pass; no expansion or post-processing is required. Latent action decoding therefore introduces zero additional computational overhead, and the token-level compression achieved by reparameterization translates directly into proportional savings in prefill computation, KV-cache memory, and end-to-end inference latency (Table~\ref{tab:efficiency}).

\textbf{Reparameterization rate.}
We quantify the degree of compression as $r = \sum_{i=1}^{N} |\hat{\tau}_i| \,/\, \sum_{i=1}^{N} |\tau_i|$, where $|\tau_i|$ and $|\hat{\tau}_i|$ are the token counts of the $i$-th original and reparameterized trajectories. A smaller $r$ corresponds to higher compression. Because the identification procedure ranks candidates by $\text{score}(s)$, segments with the strongest evidence of transition invariance are abstracted first, inducing a natural priority ordering exploited in the progressive abstraction ablation (Section~\ref{sec:progressive}).

\vspace{-0.7em}
\subsection{Applicability and Failure Modes}
\label{sec:fail}

\vspace{-0.3em}
LAR is effective when the agent's action space contains a substantial subset of executable latent actions. Tasks with rich structural scaffolding, such as web interaction with protocol-constrained tool invocations or code generation with recurring syntactic patterns, admit larger compressible subsets, whereas reasoning-intensive tasks with diverse free-form content admit smaller ones.

\textbf{Failure mode and concrete instance.}
Failure arises when abstraction merges segments that are not transition-equivalent: the latent action no longer represents a single behavior, and replacing concrete actions alters the induced transitions. As a concrete example, consider a TriviaQA trajectory where the agent must convey the query \textit{``Next British Prime Minister after Arthur Balfour''} to a search tool (the same trajectory analyzed in Section~\ref{sec:case_study} and Figure~\ref{fig:case_stude}). Under default thresholds, the query is preserved explicitly because its high next-token entropy exceeds $H_{\max}$, but if $H_{\max}$ is raised aggressively, the entropy filter may admit query-adjacent patterns and replace the query itself with a latent action. The tool interaction then breaks entirely, producing an abrupt rather than gradual performance degradation: a categorical breakdown of environment-facing transitions, characterized empirically as the Phase III collapse in our progressive abstraction ablation (Section~\ref{sec:progressive}).

\textbf{Prevention by design.}
LAR addresses this failure mode through prevention at the identification stage rather than runtime fallback. The entropy filter introduced in Section~\ref{sec:implementation} excludes high-entropy, parameter-binding segments before they enter the latent action vocabulary, while the frequency filter $\text{freq}(s) \geq f_{\min}$ excludes long-tail patterns lacking sufficient statistical support. Once an incorrect latent action is trained into the model, runtime rollback is difficult, so filtering at the identification stage eliminates the need for runtime mitigation. This makes LAR conservative by default: it improves efficiency on the common, structurally regular portions of the action space while introducing no risk on the rare, irregular portions. The method therefore does not seek maximal compression but identifies the largest subset of latent actions that preserve transition behavior; the empirical boundary of this subset is characterized in Section~\ref{sec:progressive}.

\vspace{-0.5em}
\section{Main experiment}
\label{sec:experiment}

\vspace{-0.5em}
\subsection{Experimental Setup}

\textbf{Backbone Models:} We evaluate LAR on two widely used instruction-tuned LLMs: \texttt{Meta-Llama-3.1-8B-Instruct}~\citep{dubey2024llama} and \texttt{Qwen3-8B}~\citep{yang2025qwen3}. These models allow us to assess whether action space reparameterization generalizes across model families. For each model, latent actions are learned exclusively from its own rollout trajectories.

\textbf{Benchmarks:} We consider a diverse set of LLM agent benchmarks covering different interaction patterns and action structures. TriviaQA~\citep{joshi2017triviaqa} represents multi-step reasoning tasks; KodCode~\citep{xu2025kodcode} represents code-generation tasks with highly structured action patterns; and Mind2Web~\citep{deng2023mind2web} represents web-based, tool-using tasks with rich interaction scaffolds. For code benchmarks, latent actions are learned jointly across datasets to evaluate cross-task generalization within a shared action style.

\textbf{Baselines:} We choose vanilla LLM agents and ReAct-style agents, as well as representative efficiency-oriented methods operating at different stages of the agent pipeline, including token-level generation control (TokenSkip~\citep{xia2025tokenskip}, ConciseHint~\citep{tang2025concisehint}) and context/memory optimization (ACON~\citep{kang2025acon}).

\textbf{Evaluation Metrics:} We report task-specific performance metrics (e.g., accuracy or success rate) and the relative reduction in action token counts compared to the original (Vanilla) trajectories. All methods are evaluated using identical decoding settings and hardware configurations. More experimental information is detailed in Appendix~\ref{sec:ExperimentalSetup}.

\begin{table*}[t]
\vspace{-1em}
\caption{Main results on three representative LLM agent benchmarks. We compare LAR with general prompting baselines and efficiency-oriented methods operating at the token or context level, across two backbone models. Numbers report task performance, with parentheses indicating the relative change in action tokens. Best performance for each backbone and benchmark is highlighted in \textbf{bold}.}
\centering
\begin{tabular}{llccc}
\toprule
\small
\textbf{Backbone} & \textbf{Method} & \textbf{TriviaQA} & \textbf{KodCode} & \textbf{Mind2Web} \\
\midrule
\multirow{7}{*}{Qwen3-8B}
  & Vanilla & 67.40 & 34.44 & 36.73 \\
  & COT & 69.43 & 35.10 & 34.15 \\
  & ReAct & 77.84 & 53.64 & - \\
  \cmidrule{2-5}
  & TokenSkip & 57.02 (-28.7\%) & 29.80 (-28.4\%) & 31.13 (-16.6\%) \\
  & ACON & 55.33 (-27.9\%) & 28.67 (-22.5\%) & 30.77 (-17.7\%) \\
  & ConciseHint & 68.69 (-12.7\%) & 28.47 (-12.5\%) & 35.33 (+11.9\%) \\
  \cmidrule{2-5}
  & \textbf{LAR} & \hlc{\textbf{80.09} (-27.1\%)} & \hlc{\textbf{54.30} (-9.2\%)} & \hlc{\textbf{39.84} (-2.9\%)} \\
\midrule
\multirow{7}{*}{\shortstack{Llama-3.1\\8B-Instruct}}
  & Vanilla & 73.63 & 31.13 & 24.40 \\
  & COT & \textbf{75.50} & 22.52 & 13.85 \\
  & ReAct & 59.88 & 33.11 & - \\
  \cmidrule{2-5}
  & TokenSkip & 68.87 (-12.0\%) & 26.03 (-5.8\%) & 14.27 (-1.9\%) \\
  & ACON & 57.14 (-25.4\%) & 24.67 (-23.1\%) & 15.63 (-16.7\%) \\
  & ConciseHint & 67.32 (-13.5\%) & 25.83 (-12.0\%) & 17.33 (-13.0\%) \\
  \cmidrule{2-5}
  & \textbf{LAR} & \hlc{72.46 (-23.3\%)} & \hlc{\textbf{35.10} (-9.8\%)} & \hlc{\textbf{28.30} (-20.8\%)} \\
\bottomrule
\end{tabular}
\label{tab:main_results}
\vspace{-1.5em}
\end{table*}

\vspace{-0.2em}
\subsection{Main Results: Performance and Efficiency Analysis}
\label{sec:main_results}

\vspace{-0.5em}
Table~\ref{tab:main_results} reports task performance alongside the relative reduction in action tokens (in parentheses, compared to the original Vanilla trajectories). We analyze the results from three perspectives: the accuracy-efficiency trade-off, robustness across interaction regimes and backbones, and overall comparison with baselines.

\textbf{Accuracy-Efficiency Trade-off under Action Reparameterization:} As shown in Table~\ref{tab:main_results}, LAR consistently achieves favorable accuracy-efficiency tradeoffs across backbones and benchmarks. In most settings, LAR reduces the effective decision horizon, reflected in fewer action tokens, while preserving or improving task success rates. 
These results suggest that LAR primarily eliminates redundant decision steps rather than semantically critical ones.

Unlike TokenSkip, ACON, and ConciseHint, which intervene at the token generation or conditioning stage and may destabilize decision semantics, LAR reparameterizes the action space itself into executable latent units, preserving task-relevant transition behavior.
The results also reveal abstraction boundaries. On TriviaQA with Llama-3.1-8B-Instruct, LAR achieves a 23.3\% token reduction but incurs a slight accuracy drop versus Vanilla and CoT, suggesting that when structurally redundant action segments are limited, further abstraction approaches the boundary of semantic decision-making rather than indicating instability. Overall, LAR occupies a more favorable region in accuracy-efficiency space, supporting the conclusion that the effective decision horizon, rather than token count alone, is the dominant factor governing LLM agent efficiency.

\textbf{Robustness across Heterogeneous Interaction Regimes and Backbone Behaviors:} 
Table~\ref{tab:main_results} further shows that LAR generalizes robustly across benchmarks with distinct interaction structures, reasoning-intensive retrieval (TriviaQA), structured code generation (KodCode), and tool-using web interaction (Mind2Web), where the degree of improvement is jointly determined by task-level structural regularity and backbone-specific generation behavior. On KodCode, LAR matches or closely approaches the strongest baselines while reducing the decision horizon by approximately 9 to 10\%, suggesting that code-generation tasks contain substantial executable structural redundancy amenable to abstraction. On Mind2Web, performance gains are more strongly modulated by backbone behavior: Qwen3-8B achieves the highest accuracy with modest token reduction, while Llama-3.1-8B-Instruct yields a substantial accuracy improvement and a larger reduction in decision horizon, reflecting differences in the volume of intermediate text generated and, thus, the fraction of action content that can be safely abstracted. Overall, these results indicate that action representation learning serves as a mechanism complementary to model- and system-level optimizations for efficient LLM agent inference.

\textbf{Overall Performance Comparison:} Across all benchmarks and backbones, LAR consistently achieves the best or near-best task performance among efficiency-oriented methods, while substantially outperforming existing agentic baselines, which reduce inference cost through token pruning or context compression but frequently incur severe performance degradation on structured tasks like KodCode and Mind2Web. In contrast, LAR maintains strong task performance by operating over a reparameterized latent action space, rather than applying token- or context-level compression, thereby preserving environment-facing transition semantics while eliminating structural redundancy in agent behavior. These results support our central claim that enforcing executability constraints during reparameterization is critical for achieving favorable efficiency-performance trade-offs without sacrificing task success.

\begin{wrapfigure}{r}{0.5\textwidth}
    \centering
    \vspace{-1.2em}
    \captionof{table}{Held-out benchmark generalization result. LAR is compared against ReAct under identical backbone settings.}
    \footnotesize
    \setlength{\tabcolsep}{4pt}
    \begin{tabular}{llccc}
    \toprule
    \textbf{Backbone} & \textbf{Method} & \textbf{Musique} & \textbf{HumanEval} & \textbf{MBPP} \\
    \midrule
    \multirow{2}{*}{Qwen3-8B}
      & ReAct & 27.61 & 89.63 & 74.17 \\
      & LAR & 26.57 & 91.46 & 75.50 \\
    \midrule
    \multirow{2}{*}{\shortstack{Llama-3.1\\8B-Instruct}}
      & ReAct & 20.93 & 56.71 & 48.40 \\
      & LAR & 22.35 & 60.37 & 46.60 \\
    \bottomrule
    \end{tabular}
    \label{tab:cross_comparison}
    \vspace{-1em}
\end{wrapfigure}

\vspace{-0.7em}
\subsection{Held-out Benchmark Generalization of Latent Actions}
\label{sec:cross_benchmark}

\vspace{-0.3em}
This experiment evaluates whether latent actions learned by LAR capture reusable decision structure or merely encode dataset-specific artifacts. We first introduce another three datasets: Musique~\cite{trivedi2022musique} for QA tasks, MBPP~\cite{austin2021program} and HumanEval~\cite{chen2021evaluating} for coding tasks. We regard them as a held-out benchmark because we do not use them for any retraining or adaptation. Corresponding to this, we regard TriviaQA and KodCode as held-in datasets. Specifically, we train latent actions using trajectories collected on held-in benchmarks and directly apply the resulting action reparameterization to other held-out benchmarks.

This setup directly tests the core assumption underlying LAR's design.
If latent actions correspond to transition-equivalent and executable decision units (Section~\ref{sec:method}), they should generalize across benchmarks that share a common action domain even when surface distributions differ. Conversely, if they merely overfit to dataset-specific patterns, their effectiveness should deteriorate when transferred to unseen benchmarks.

As shown in Table~\ref{tab:cross_comparison}, LAR demonstrates strong held-out benchmark generalization across both backbones. When trained on a held-in dataset and evaluated on a held-out benchmark, LAR consistently matches or outperforms ReAct without task-specific retraining or prompt engineering. This behavior indicates that LAR learns domain-level action abstractions for tasks, rather than benchmark-specific heuristics. In particular, the latent actions appear to encode stable structural behaviors such as code scaffolding and formatting patterns that remain valid across datasets.

\section{Mechanism Analysis and Case Study}

\vspace{-0.7em}
\subsection{Action Equivalence Analysis on Latent Action Reparameterization~\label{sec:equivalence}}

\vspace{-0.5em}
LAR achieves performance gains with efficient action reparameterization. However, two questions remain: \textit{whether the reparameterized latent action is functionally equivalent to the original action sequence}, and \textit{whether the observed performance improvements can be mechanistically attributed to the abstraction induced by reparameterization}. To eliminate the confounding effect of sequence length reduction at inference, we manually append padding tokens following the reparameterized sequence to match the length of the original action sequence.

\begin{wrapfigure}{r}{0.55\textwidth}
    \vspace{-1.2em}
    \captionof{table}{Action equivalence of LAR. LAR-PT denotes LAR inference with padding tokens.}
    \centering
    \footnotesize
    \setlength{\tabcolsep}{4pt}
    \begin{tabular}{llccc}
        \toprule
        \textbf{Backbone} & \textbf{Method} & \textbf{TriviaQA} & \textbf{KodCode} & \textbf{Mind2Web} \\
        \midrule
        \multirow{3}{*}{Qwen3-8B}
          & ReAct & 77.84 & 53.64 & 34.15 \\
          & LAR & 80.09 & 54.30 & 39.84 \\
          & LAR-PT & 79.64 & 53.64 & 34.40 \\
        \midrule
        \multirow{3}{*}{\shortstack{Llama-3.1\\8B-Instruct}}
          & ReAct & 59.88 & 33.11 & 24.40 \\
          & LAR & 72.46 & 35.10 & 28.30 \\
          & LAR-PT & 74.25 & 34.44 & 27.15\\
        \bottomrule
    \end{tabular}
    \vspace{-1.7em}
    \label{tab:equality}
\end{wrapfigure}

Table~\ref{tab:equality} shows the result. LAR-PT performs better than ReAct across all settings, demonstrating that the language model derives performance gains uniquely from action abstraction, independent of sequence length reduction. Nevertheless, LAR still marginally outperforms LAR-PT, particularly on Mind2Web, indicating that inference efficiency from sequence compression provides an additional, complementary source of performance improvement. 

\begin{wrapfigure}{r}{0.6\textwidth}
    \centering
    \vspace{-1.2em}
    \begin{minipage}{0.295\textwidth}
        \centering
        \includegraphics[width=\textwidth]{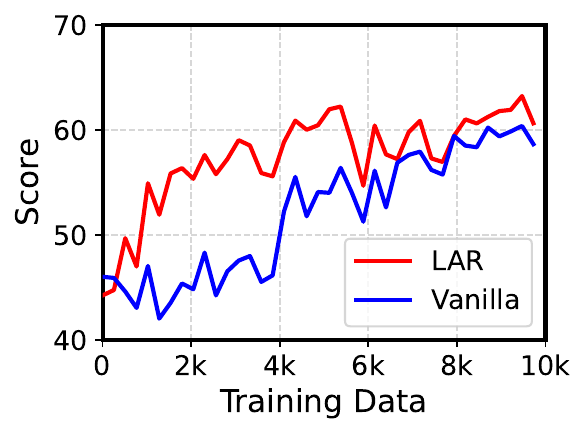}
        \caption*{\small (a) TriviaQA with Llama}
    \end{minipage}
    \hfill
    \begin{minipage}{0.295\textwidth}
        \centering
        \includegraphics[width=\textwidth]{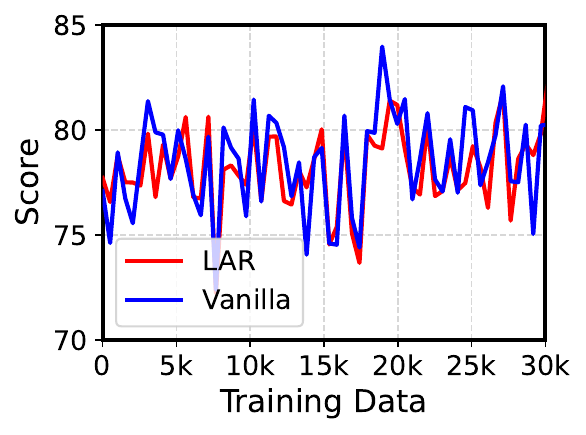}
        \caption*{\small (b) TriviaQA with Qwen3}
    \end{minipage}
    \caption{These two curves indicate two different characteristics of LAR on further policy optimization: (a) converges faster; (b) consistent learning stability.}
    \label{fig:triviaqa}
    \vspace{-1.4em}
\end{wrapfigure}

\vspace{-0.7em}
\subsection{Learning Stability of Latent Action Reparameterization~\label{sec:learning_stablity}}

\vspace{-0.5em}
To validate the hypothesis that action reparameterization induces more consistent and stable rollout trajectories, which in turn leads to improved learning stability during policy optimization, we apply GRPO under both the vanilla backbone model and the post-LAR training model and compare the convergence behavior of the training score curves.
The detailed experimental settings are in Appendix~\ref{app:reproducibility}.

Figure~\ref{fig:triviaqa} shows the score curves on these two models. LAR generally converges faster and exhibits nearly the same stability during the GRPO training process. It demonstrates that reduced rollout variance induced by reparameterization leads to more homogeneous trajectory distributions, thereby providing a more stable learning signal and enabling faster convergence of the policy optimization process.

\vspace{-0.7em}
\subsection{Progressive Abstraction Ablation and the Boundary of Executable Latent Actions}
\label{sec:progressive}


\vspace{-0.5em}
This experiment investigates the question: where the abstraction boundary lies for latent action reparameterization, through a structured ablation over abstraction strength. While previous results show that moderate abstraction can improve efficiency without harming performance, we conduct a progressive abstraction ablation to characterize both the beneficial regime and the failure mode of LAR by gradually increasing the reparameterization rate and observing when task performance begins to degrade. This analysis directly corresponds to the failure modes discussed in Section~\ref{sec:fail}.

\textbf{Experimental Design (Progressive Abstraction Ablation):}
We perform a controlled ablation by progressively increasing the reparameterization rate, defined as the proportion of action segments replaced by latent actions, while keeping the backbone model and decoding settings fixed. This ablation systematically varies the degree of abstraction applied to the agent's action space, allowing us to trace how performance evolves as abstraction moves from low-entropy structural components toward high-entropy parameterized components.

\textbf{Phase I: Moderate Abstraction (Ablation Regime):}

\begin{wrapfigure}{r}{0.63\textwidth}
    \centering
    \begin{minipage}[t]{0.31\textwidth}
        \centering
        \includegraphics[width=\linewidth]{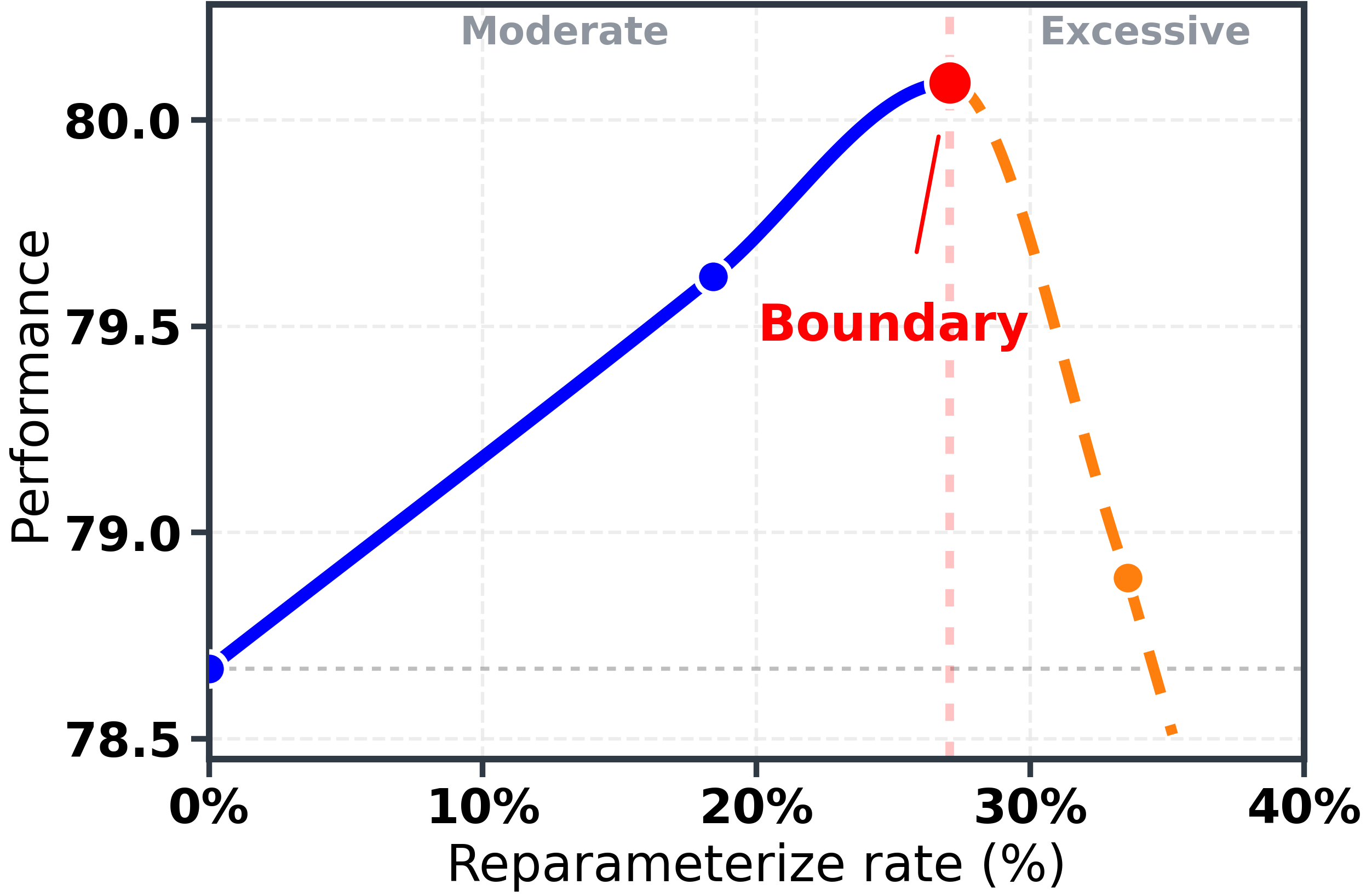}
        \caption*{(a) TriviaQA (Qwen3-8B)}
    \end{minipage}
    \hfill
    \begin{minipage}[t]{0.31\textwidth}
        \centering
        \includegraphics[width=\linewidth]{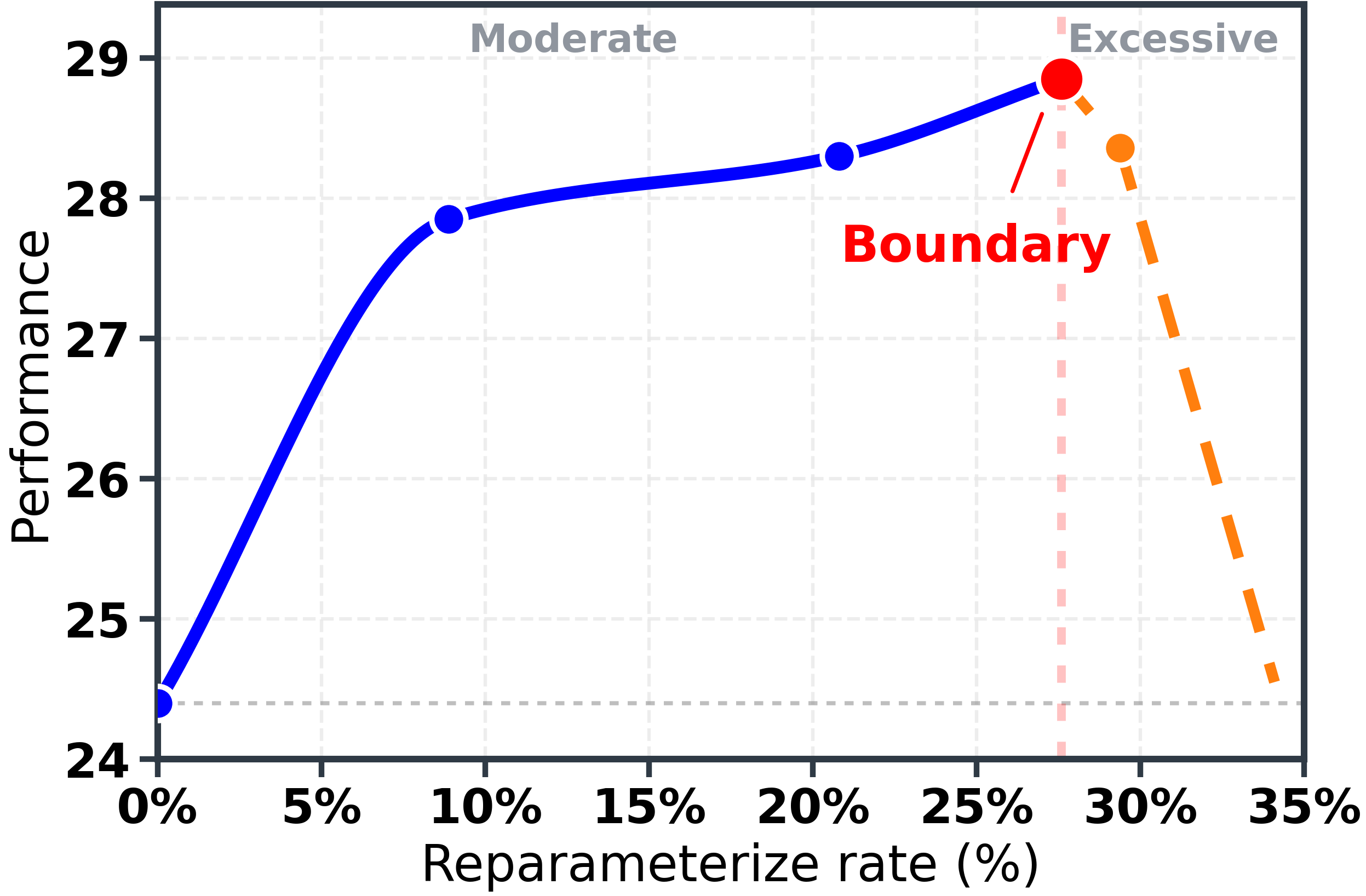}
        \caption*{(b) Mind2Web (Llama-3.1-8B)}
    \end{minipage}
    \caption{
    Progressive abstraction ablation results. Task performance as a function of the reparameterization rate for two representative settings. Moderate abstraction improves performance by eliminating low-entropy structural redundancy, while excessive abstraction leads to a sharp performance collapse once high-entropy, parameterized components are abstracted.
    }
    \label{fig:ablation_boundary}
    \vspace{-1.5em}
\end{wrapfigure}

In the low-to-moderate abstraction regime of this ablation, performance consistently improves alongside inference efficiency. This regime corresponds to the selective abstraction of low-entropy structural components, such as recurring scaffolds, protocol formats, and stable interaction patterns. Removing these redundant decision steps shortens the effective decision horizon and reduces inference cost, while preserving task-relevant semantics. The observed performance gains in this ablation regime indicate that moderate abstraction removes redundant or noisy action fragments rather than eliminating essential decisions. This trend is visually illustrated in Fig.~\ref{fig:ablation_boundary}, which shows a consistent performance increase as low-entropy structural components are progressively abstracted.

\textbf{Phase II: Abstraction Boundary:}
As the reparameterization rate continues to increase, performance peaks, marking the abstraction boundary. At this point, most transition-invariant, executable components have been safely abstracted and further abstraction yields no additional benefit. This peak empirically delineates the maximal scope of abstraction that preserves transition equivalence and executability.

\textbf{Phase III: Performance Collapse:}
Beyond the abstraction boundary, performance degrades sharply. This collapse occurs when the ablation begins to include high-entropy parameterized components, such as queries, entity references, or task-specific arguments. Abstracting these components violates the executability condition, as latent actions no longer correspond to transition-equivalent behaviors across contexts. Notably, the degradation is abrupt rather than gradual, reflecting a semantic failure mode caused by broken environment-facing transitions rather than insufficient modeling capacity.

\textbf{Cross-Task Consistency:}
The three-phase behavior along the abstraction ablation axis—performance improvement, peak, and collapse—appears consistently across both Mind2Web and TriviaQA. While the exact location of the abstraction boundary varies with task and backbone, the qualitative trend is shared across domains, suggesting that the abstraction boundary is a structural property of action reparameterization rather than an artifact of a specific task.

\textbf{Method-Level Takeaway:}
This ablation provides direct empirical support for LAR's design choice to restrict abstraction to executable latent actions. Rather than aiming for maximal compression, LAR identifies the largest subset of low-entropy, transition-invariant action segments that can be safely abstracted. The observed performance collapse beyond the abstraction boundary validates this restriction and highlights why executability fundamentally constrains useful action representations.

\vspace{-0.8em}
\subsection{Case study}
\label{sec:case_study}

\vspace{-0.5em}
To better reveal the characteristics of LAR, we conduct a case analysis on a multi-step reasoning instance from TriviaQA to illustrate how LAR reshapes the agent's action structure. As shown in Fig.~\ref{fig:case_stude}, the vanilla agent generates a long sequence of fine-grained textual actions for information retrieval and answer construction, many of which correspond to recurrent structural scaffolds such as reasoning templates and tool invocation formats, rather than task-critical semantic decisions.

Under LAR, this trajectory is reformulated into two classes of actions: executable latent actions and explicit high-entropy parameterized components. Low-entropy, recurrent structural patterns are abstracted into latent actions representing transition-equivalent behaviors, while parameter-binding elements that determine task semantics, including the concrete search query and final answer, remain explicit to ensure executability.

This case study highlights that LAR shortens the effective decision horizon by abstracting transition-invariant structure, while preserving correct interaction with external tools by retaining parameter-rich components. Together, these effects illustrate how selective action reparameterization improves inference efficiency without compromising semantic correctness or executability.

\vspace{-0.8em}
\section{Conclusion}

\vspace{-0.5em}
We introduced Latent Action Reparameterization (LAR), a principled framework that improves LLM-agent efficiency by treating \emph{action representation} as a first-class modeling choice. By learning executable latent actions that compress recurrent low-entropy structures while preserving high-entropy, parameter-binding content, LAR redefines the unit of decision making and directly addresses the inefficiency caused by overly fine-grained action interfaces, achieving favorable performance--efficiency trade-offs across diverse tasks and backbone models.

\bibliography{neurips_2026}
\bibliographystyle{plainnat}

\clearpage

\appendix

\section{Detailed Experimental Setup and Design Rationale}
\label{sec:ExperimentalSetup}

\subsection{Agent Models and Training Protocol}
\label{app:agent-models}

We conduct experiments using \textbf{Meta-Llama-3.1-8B-Instruct}~\citep{dubey2024llama} and \textbf{Qwen3-8B}~\citep{yang2025qwen3} to ensure that observed efficiency gains are not specific to a single model family. No architectural modifications are applied to the base models: the attention mechanisms, layer counts, and hidden dimensions remain unchanged. To integrate latent action symbols into the model, we employ \emph{parameter-efficient adaptation} via LoRA (rank $r=8$, scaling factor $\alpha=16$, applied a to the query, key, value, and output projection matrices) together with newly added embedding and output-head entries for the latent action vocabulary $\mathcal{Z}$. Only the LoRA adapter weights and the new latent action embeddings are trainable, amounting to approximately $0.1\%$ of total model parameters; full optimization details are summarized in Appendix~\ref{app:reproducibility}. All original pretrained weights, including all pre-existing token embeddings, remain frozen throughout training. This is parameter-efficient adaptation rather than full fine-tuning, and the original base model is preserved without modification.

Latent action vocabularies are constructed separately for each model using trajectories generated by the same model. This ensures that reparameterization does not rely on cross-model transfer of action representations and that the learned latent actions reflect the generation behavior of the specific backbone. Concrete configurations and the resulting vocabulary sizes for each (model, benchmark) pair are reported in Appendix~\ref{app:lar-config}.

\renewcommand{\algorithmiccomment}[1]{\quad$\triangleright$ #1}

\subsection{Latent Action Identification Algorithm}
\label{app:algorithm}

Algorithm~\ref{alg:identify} provides the full segment identification procedure summarized in the main text. The algorithm extracts word-level $n$-grams within sentence boundaries, applies frequency and entropy filters that operationalize the transition equivalence condition, ranks candidates by a composite score that balances coverage and behavioral stability, and removes redundant entries so that each retained latent action is maximally informative.

\begin{algorithm}[h]
\caption{Latent Action Identification}
\label{alg:identify}
\begin{algorithmic}[1]
\REQUIRE Trajectory dataset $\mathcal{D} = \{\tau_1, \dots, \tau_N\}$; minimum frequency $f_{\min}$; maximum entropy $H_{\max}$; $n$-gram size range $[n_{\text{lo}}, n_{\text{hi}}]$; capacity $K$; overlap threshold $\rho$
\ENSURE Latent action set $\mathcal{Z}$
\STATE $\mathcal{C} \leftarrow \emptyset$ \COMMENT{candidate segments}
\FOR{each trajectory $\tau \in \mathcal{D}$}
    \STATE Extract all word-level $n$-grams of size $n \in [n_{\text{lo}}, n_{\text{hi}}]$ within sentence boundaries
    \STATE $\mathcal{C} \leftarrow \mathcal{C} \cup \{\text{extracted } n\text{-grams}\}$
\ENDFOR
\STATE $\mathcal{C}_{\text{freq}} \leftarrow \{ s \in \mathcal{C} \mid \text{freq}(s) \geq f_{\min} \}$ \COMMENT{frequency filter}
\FOR{each $s \in \mathcal{C}_{\text{freq}}$}
    \STATE Compute the next-token entropy $H(s)$ as defined in the main text
\ENDFOR
\STATE $\mathcal{C}_{\text{ent}} \leftarrow \{ s \in \mathcal{C}_{\text{freq}} \mid H(s) \leq H_{\max} \}$ \COMMENT{entropy filter}
\STATE Rank $\mathcal{C}_{\text{ent}}$ by $\text{score}(s) = \text{freq}(s) / (H(s) + 1)$ in descending order
\STATE $\mathcal{Z} \leftarrow \emptyset$
\FOR{$s$ in ranked $\mathcal{C}_{\text{ent}}$}
    \IF{$s$ is not a substring of any $s' \in \mathcal{Z}$ \AND $\text{overlap}(s, s') < \rho$ for all $s' \in \mathcal{Z}$}
        \STATE $\mathcal{Z} \leftarrow \mathcal{Z} \cup \{s\}$
    \ENDIF
    \IF{$|\mathcal{Z}| = K$}
        \STATE \textbf{break}
    \ENDIF
\ENDFOR
\RETURN $\mathcal{Z}$
\end{algorithmic}
\end{algorithm}

We use the overlap threshold $\rho = 0.7$ across all benchmarks. Per-benchmark settings of $f_{\min}$, $H_{\max}$, $[n_{\text{lo}}, n_{\text{hi}}]$, and $K$ are reported in Appendix~\ref{app:lar-config}.

\subsection{Reproducibility and Implementation Details}
\label{app:reproducibility}

\textbf{Hardware and decoding.}
All models and baselines are trained and evaluated on servers equipped with $8\times$H200 140GB GPUs. We use vLLM for inference with temperature $T=0$ for deterministic decoding. All experiments are conducted with fixed random seeds.

\textbf{LAR Training configuration.}
LAR training follows the trajectory-level distillation procedure described in Section~\ref{sec:implementation}, with the parameter-efficient adaptation setup (LoRA plus newly added latent action embeddings, all original weights frozen) summarized in Appendix~\ref{app:agent-models}. The full optimization configuration is summarized in Table~\ref{tab:training-config}. We use a pure KL distillation objective ($\lambda = 1.0$) computed over shared content positions $M$, with distillation temperature $\tau = 2.0$.

\begin{table}[h]
\centering
\small
\caption{LAR training configuration.}
\label{tab:training-config}
\begin{tabular}{ll}
\toprule
\textbf{Component} & \textbf{Value} \\
\midrule
Optimizer & AdamW \\
Effective batch size & 16 (via gradient accumulation) \\
Learning rate & $1 \times 10^{-4}$ \\
LR schedule & Cosine \\
Warmup steps & 100 \\
Weight decay & 0.01 \\
Epochs & 3 \\
Precision & fp16 \\
Distillation temperature $\tau$ & 2.0 \\
KL weight $\lambda$ & 1.0 \\
LoRA rank $r$ & 8 \\
LoRA scaling $\alpha$ & 16 \\
LoRA target modules & q, k, v, o projections \\
Hardware & $8\times$ H200 140GB GPUs \\
\bottomrule
\end{tabular}
\end{table}

\begin{table}[t]
\centering
\small
\setlength{\tabcolsep}{5pt}
\caption{Main training hyperparameters for GRPO on TriviaQA and KodCode.}
\begin{tabular}{lcc}
\toprule
Parameter & TriviaQA & KodCode\\
\midrule
Train batch size & 256 & 512 \\
Validation batch size & 128 & 128 \\
Learning rate & $1\times10^{-6}$ & $1\times10^{-6}$ \\
LR warmup ratio & 0.285 & 0.285 \\
Max prompt length & 4096 & 4096 \\
Max response length & 1024 & 8192 \\
Max observation length & 1024 & 2048 \\
Rollout agents & 4 & 8 \\
Max turns & 4 & 10 \\
KL loss coef & $0.001$ & $0.001$ \\
\bottomrule
\end{tabular}
\label{tab:grpo_training}
\end{table}

\textbf{Trajectory data.}
Latent actions are identified from agent rollout trajectories generated by each backbone on the corresponding benchmark training split. The amount of trajectory data used for identification and distillation is benchmark-dependent; for example, on KodCode, we sample 20K trajectories from the full 447K dataset for post-training, balancing identification quality against compute cost. Per-benchmark data sizes and corresponding latent action vocabulary configurations are reported in Appendix~\ref{app:lar-config}.

\textbf{GRPO Training Setting for Learning Stability Analysis.} The code for GRPO training is adapted from Search-R1~\footnote{\url{https://github.com/PeterGriffinJin/Search-R1}}, which is established on VeRL framework~\cite{sheng2024hybridflow}. We conduct experiments on TriviaQA and KodCode to support the learning stability analysis in Section~\ref{sec:learning_stablity}. The detailed training settings are in Table~\ref{tab:grpo_training}.

\textbf{Source code.}
Complete source code, including the segment identification pipeline, training scripts, and evaluation harness, is publicly released at the anonymous repository linked in the abstract.

\subsection{Per-Task LAR Configuration}
\label{app:lar-config}

The latent action identification pipeline (Algorithm~\ref{alg:identify}) is configured per benchmark to reflect differences in action structure: tasks with rich, recurring interaction scaffolds admit smaller frequency thresholds and larger compressible fractions, whereas reasoning-intensive tasks with diverse free-form content require stricter frequency thresholds to ensure that only stable patterns are abstracted. Table~\ref{tab:lar-config} summarizes the configuration used for each benchmark in our experiments.

\begin{table}[h]
\centering
\small
\caption{Per-benchmark LAR configuration. $f_{\min}$ is the minimum frequency threshold; $H_{\max}$ is the maximum next-token entropy threshold; $[n_{\text{lo}}, n_{\text{hi}}]$ denotes the $n$-gram size range; $K$ is the latent action capacity. ``\#Latent actions'' reports the resulting vocabulary size after frequency, entropy, and redundancy filters; ``Avg.\ words/action'' reports the average segment length across the retained latent actions.}
\label{tab:lar-config}
\begin{tabular}{lccccc}
\toprule
\textbf{Benchmark} & \textbf{$n$-gram range} & \textbf{$f_{\min}$} & \textbf{$H_{\max}$} & \textbf{\# Latent actions} & \textbf{Avg.\ words/action} \\
\midrule
TriviaQA  & $[3, 5]$ & 2000 & 10.0 & 200--1000 & 3--5 \\
KodCode   & $[2, 6]$ & 10   & 10.0 & $\sim$100 & 2--6 \\
Mind2Web  & $[2, 6]$ & 1000 & 10.0 & $\sim$100 & 2--6 \\
\bottomrule
\end{tabular}
\end{table}

\textbf{Configuration rationale.}
The per-benchmark differences reflect characteristics of the action structure observed in each domain. TriviaQA combines free-form multi-step reasoning with retrieval calls, producing many recurring reasoning templates and tool-invocation scaffolds; we therefore use a wider $n$-gram range (3--5) and a high frequency threshold ($f_{\min}=2000$) to retain only stable, broadly applicable patterns. KodCode contains highly structured code-generation trajectories with consistent syntactic and protocol scaffolds shared across diverse problems; the large pool of stable patterns allows a lower frequency threshold ($f_{\min}=10$) while still ensuring statistical reliability, and the resulting $\sim$100 latent actions transfer directly to HumanEval and MBPP without further identification, as reported in Section~\ref{sec:cross_benchmark}. Mind2Web involves protocol-constrained tool invocations and substantial HTML scaffolding; we additionally extract recurring HTML tag sequences alongside textual $n$-grams to capture the protocol-level repetitions characteristic of web interaction.

The redundancy removal threshold $\rho = 0.7$ is held constant across benchmarks, as is the entropy threshold $H_{\max} = 10.0$. The latter is set conservatively to ensure that high-entropy parameter-binding content (search queries, entity names, code identifiers) is reliably preserved in the explicit output space.

\textbf{Trajectory data sizes.}
For latent action identification (Appendix~\ref{app:algorithm}) and distillation (Appendix~\ref{app:reproducibility}), we sample agent rollout trajectories from each benchmark's training split. Concretely, we use 20K trajectories for Koemphasize of the 447K full set), and analogous sampling for TriviaQA and Mind2Web (sized to provide reliable frequency and entropy estimates while keeping post-training cost modest). HumanEval and MBPP are evaluated in a strict zero-shot transfer setting using the latent action vocabulary learned exclusively from KodCode.

\subsection{Benchmark Selection and Task Categorization}
Benchmarks are selected to span distinct agent behaviors and action structures. QA benchmarks (\textbf{TriviaQA}, \textbf{Musique}) emphasizes multi-step reasoning with relatively low structural repetition. Code benchmarks (\textbf{HumanEval}, \textbf{MBPP}, \textbf{KodCode}) exhibit strong syntactic regularities and recurring generation patterns, making them suitable for studying reusable semantic action units. \textbf{Mind2Web} involves complex tool interactions with extensive protocol-level scaffolds and repeated system-level configurations, providing a setting where executable action abstraction is particularly impactful.

\subsection{Latent Action Learning for Held-out Benchmarks: Source and Transfer Setting}
\label{app:code-tasks}

We evaluate the generalization of LAR on two domains of tasks. For the QA domain, latent actions are identified and trained from TriviaQA trajectories alone, while for the coding domain, latent actions are identified and trained from KodCode trajectories, following the procedure in Appendix~\ref{app:algorithm} and Appendix~\ref{app:reproducibility}. The resulting vocabulary and the trained embeddings are then directly applied to Musique (HumanEval and MBPP for code) without any additional retraining, adaptation, or prompt engineering. This corresponds to a held-out benchmark transfer setting, in which the latent actions are never exposed to the evaluated benchmark.

This setup is designed to test whether latent actions capture domain-level structural regularities, including function scaffolds, formatting conventions, and invocation patterns shared across different tasks, rather than dataset-specific artifacts. The corresponding empirical results are reported in our held-out benchmark generalization experiments.

\subsection{Baseline Methods}
Baselines are chosen to represent distinct efficiency paradigms:
\begin{itemize}
    \item \textbf{Vanilla LLM agents}, which operate directly over token-level actions;
    \item \textbf{ReAct-style agents}, which interleave reasoning and acting;
    \item \textbf{Token-level efficiency methods}, which regulate generation dynamics (TokenSkip, ConciseHint);
    \item \textbf{Context and memory optimization methods}, which compress interaction histories (ACON).
\end{itemize}
All baselines preserve the original decision interface and do not alter the action space representation.

\subsection{Baseline Evaluation}
\begin{itemize}
    \item \textbf{Vanilla LLM agents}, which operate directly over token-level actions;
    \item \textbf{ReAct-style agents}, which interleave reasoning and acting;
    \item \textbf{TokenSkip} is adapted to the COT prompt template of ours, and the cutoff lengths for LoRA adapter training are set to 4096 for \textbf{TriviaQA} and \textbf{Mind2Web}, and 8192 for \textbf{KodCode}; the compression ratio is set to 0.7, while the other settings are left unchanged.
    \item \textbf{ACON} was modified to use Qwen3-8B as both the compressor and generator, where we set the maximum generated tokens to 8192, and we tested $n$ cases for the benchmarks.
    \item \textbf{Context and memory optimization methods}, which compress interaction histories (ACON).
\end{itemize}

\begin{table*}[t]
\caption{The generalization experiment on LAR. \textbf{LAR-U} denotes the unified model trained on the dataset including trajectories from three domains. Numbers report task performance, with parentheses indicating the relative
change in action tokens.}
\centering
\begin{tabular}{llccc}
\toprule
\textbf{Backbone} & \textbf{Method} & \textbf{TriviaQA} & \textbf{KodCode} & \textbf{Mind2Web} \\
\midrule
\multirow{2}{*}{Qwen3-8B}
  & \textbf{LAR} & 80.09 (-27.3\%) & 58.28 (-9.2\%) & 39.84 (-2.9\%) \\
  & \textbf{LAR-U} & 76.05 (-12.3\%) & 62.91 (-4.1\%) & 31.18 (+11.5\%) \\
\midrule
\multirow{2}{*}{\shortstack{Llama-3.1\\8B-Instruct}}
  & \textbf{LAR} & 72.46 (-23.3\%) & 35.10 (-9.8\%) & 28.30 (-20.8\%) \\
  & \textbf{LAR-U} & 76.05 (-24.3\%) & 38.41 (-3.8\%) & 26.89 (-18.1\%) \\
\bottomrule
\end{tabular}
\label{tab:generalization}
\end{table*}

\subsection{Generalization across different domains~\label{app:generalization}}

To demonstrate LAR's generalizability across different domains, we merge all trajectories collected from TriviaQA, KodCode, and Mind2Web into a unified training corpus and fine-tune a single model jointly across all three domains without any domain-specific adaptation. The unified model is then evaluated on each benchmark independently to assess whether a single set of reparameterization tokens can effectively compress and represent action sequences across heterogeneous task distributions.

Table~\ref{tab:generalization} shows that our method maintains comparable performance across different tasks and domains. This result can be explained from two perspectives. On one hand, combining trajectories from diverse domains exposes the model to a richer set of reasoning patterns, thereby enhancing its general problem-solving capability. This effect is particularly pronounced for the Llama model, which benefits more substantially from the increased data diversity to strengthen its underlying reasoning ability. On the other hand, certain tasks, such as Mind2Web exhibit a moderate performance degradation under the unified setting, which can be attributed to the fact that the reparameterization tokens, when trained across heterogeneous domains, may fail to capture domain-specific structural patterns, leading to a reduced compression effectiveness and a less compact latent action representation for domain-specialized tasks.

Besides, we notice that the compression rate in five of the six settings decreases, which suggests that the unified training setting generally leads to a less aggressive compression compared to domain-specific LAR models. This is expected, as the reparameterization tokens must now encode a more heterogeneous action space spanning across QA, coding, and web navigation tasks. When trained on a single domain, the reparameterization tokens can specialize in capturing the recurring structural patterns and action primitives specific to that domain, enabling a more compact and efficient latent representation. In contrast, the unified model must learn a shared token space that accommodates the diverse action vocabularies of all three domains simultaneously, which inevitably dilutes the domain-specific compression signal and results in a more conservative encoding strategy.

\subsection{Scalability of Latent Action Reparameterization}\label{app:scalability}

To further demonstrate the scalability of our proposed LAR, so that it can easily be adopted in a larger model, we transfer the whole pipeline to Qwen3-32B~\cite{yang2025qwen3} with TriviaQA. Compared with the original LAR training setting in Table~\ref{tab:training-config}, we set LoRA rank $r=16$, LoRA scaling $\alpha=32$, learning rate $lr=5\times10^{-5}$, and training epoch $E=2$.

On testing, Qwen3-32B with ReAct framework has an accuracy of $73.20\%$, while Qwen3-32B with LAR achieves an accuracy of $75.26\%$. 
The experimental results demonstrate that LAR scales effectively to larger models, yielding a consistent performance improvement over the ReAct baseline. This gain is comparable in magnitude to those observed in 7B-model settings, suggesting that the benefit of action reparameterization is robust across model scales rather than being confined to a particular parameter regime. These findings indicate that LAR constitutes a general and model-agnostic framework, whose advantages persist as model capacity increases, underscoring its practical applicability in large-scale deployment scenarios.

\begin{table}[t]
\caption{System-level efficiency measurements on three benchmarks. \textbf{TT}: Token Throughput (tokens/s); \textbf{AGU}: Avg GPU Utilization (\%), \textbf{PG}: Peak GPU Memory (GB).}
\centering
\footnotesize
\begin{tabular}{llccccccccc}
\toprule
\multirow{2}{*}{\textbf{Model}} & \multirow{2}{*}{\textbf{Method}} 
& \multicolumn{3}{c}{\textbf{TriviaQA}} 
& \multicolumn{3}{c}{\textbf{KodCode}} 
& \multicolumn{3}{c}{\textbf{Mind2Web}} \\
\cmidrule(lr){3-5} \cmidrule(lr){6-8} \cmidrule(lr){9-11}
& & TT & AGU & PG & TT & AGU & PG & TT & AGU & PG \\
\midrule
\multirow{2}{*}{Qwen}
  & ReAct & 127.8 & 71.1 & 75.3 & 144.5 & 92.9 & 69.6 & 152.4 & 83.6 & 51.2 \\
  & LAR   & 150.2 & 58.5 & 73.1 & 146.1 & 92.5 & 69.9 & 153.9 & 80.2 & 47.7 \\
\midrule
\multirow{2}{*}{Llama}
  & ReAct & 151.6 & 56.0 & 42.5 & 140.7 & 70.7 & 70.1 & 119.0 & 16.7 & 69.8 \\
  & LAR   & 152.2 & 54.3 & 42.4 & 151.3 & 69.9 & 70.7 & 121.4 & 11.9 & 69.4 \\
\bottomrule
\end{tabular}
\label{tab:efficiency}
\end{table}

\subsection{Metrics and Evaluation Protocol}
In addition to task performance, we measure efficiency at both the decision and system levels. The \textbf{effective decision horizon} is defined as the total number of explicit generation decisions, aligning with the formal definition in Section~\ref{sec:method}. Wall-clock inference time is measured under identical hardware, decoding parameters, and maximum context lengths for all methods. Token reduction is reported as an outcome of action reparameterization rather than an explicit optimization objective.

Table~\ref{tab:efficiency} shows the three measurements related to inference cost in the hardware aspect. It shows that LAR not only accelerates the token throughput but also slightly reduces the GPU memory usage. This is because latent action decoding introduces zero additional computational overhead at inference time: latent action symbols are standard vocabulary tokens processed through the same embedding lookup and transformer forward pass as any other token.

\begin{table}[t]
\caption{The experimental results for the three Mind2Web sub-test sets.}
\centering
\small
\setlength{\tabcolsep}{4pt}
\begin{tabular}{llccc}
\toprule
\textbf{Backbone} & \textbf{Method} & \textbf{Cross-Task} & \textbf{Cross-Website} & \textbf{Cross-Domain} \\
\midrule
\multirow{2}{*}{Qwen3-8B}
  & Vanilla & 39.56 & 17.58 & 30.69 \\
  & LAR & \textbf{45.05} & \textbf{26.37} & \textbf{35.64} \\
\midrule
\multirow{2}{*}{\shortstack{Llama-3.1\\8B-Instruct}}
  & Vanilla & 22.76 & 8.13 & 22.44 \\
  & LAR & \textbf{27.64} & \textbf{11.38} & \textbf{26.69} \\
\bottomrule
\end{tabular}
\label{tab:mind2web_cross}
\end{table}

\subsection{Detailed experimental result of Mind2Web}
Mind2Web provides a unique opportunity to evaluate generalizability at three different levels: cross domains, cross websites, and cross tasks. The experimental result is shown in Table~\ref{tab:mind2web_cross}, which shows that LAR can effectively compress redundant HTML context and help LLM-based agents generate more precise actions.

\begin{figure*}[!t]
\centering
\includegraphics[width=\linewidth]{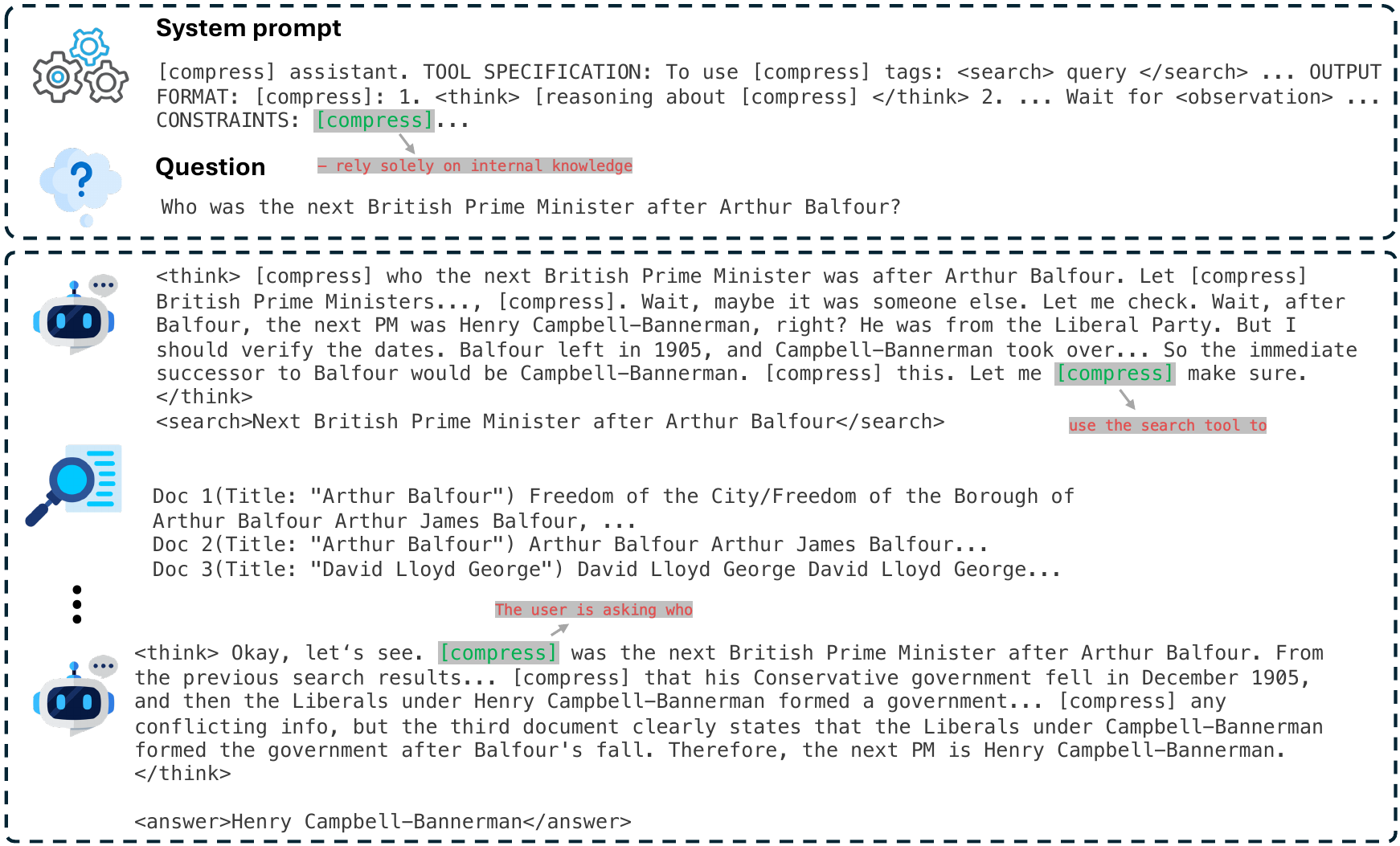}
\caption{
Case analysis of LAR on a TriviaQA example. LAR abstracts low-entropy structural components into executable latent actions while preserving high-entropy parameterized content (e.g., the search query), reducing the effective decision horizon without altering task execution.
}
\label{fig:case_stude}
\end{figure*}

\subsection{Detailed Case Analysis and Trajectory Example}
\label{app:case_analysis}
Figure~\ref{fig:case_stude} displays a complete trajectory on a TriviaQA task. The task requires the agent to identify a famous person based on a description. The Vanilla agent follows a standard ReAct-style approach, generating a sequence of thoughts, tool actions, and a final answer. This process involves generating a large number of tokens, many of which are structural and serve only to format the interaction.

LAR reparameterizes this trajectory by identifying and abstracting these recurrent structural patterns into latent actions. As shown in the figure, the lengthy sequence of tokens corresponding to the search action is compressed into a single latent token. Crucially, the high-entropy content, the search query ``Next British Prime Minister after Arthur Balfour'', is preserved explicitly to maintain executability.

This transformation results in a significantly shorter effective decision horizon. The Vanilla trajectory requires many steps to express the intent, whereas the LAR trajectory expresses the same high-level decisions in far fewer steps. By operating over these latent actions, the agent can plan and execute at a higher level of abstraction, reducing computational cost while preserving the integrity of the interaction with the environment and the final output. This example highlights how LAR selectively compresses structural redundancy while maintaining the necessary granularity for effective task performance.

\subsection{Transferability to Industrial-Grade Agent Frameworks: A Case Study on OpenClaw}
\label{app:Openclaw}

Our main benchmarks (TriviaQA, KodCode, Mind2Web) are designed for controlled scientific evaluation, but real-world LLM agents are typically deployed through industrial-grade frameworks such as OpenClaw, LangChain, and Claude Code. These frameworks embed extensive static scaffolding (tool specifications, output-format constraints, role descriptions, recurring protocol templates) into their system prompts, exhibiting exactly the structural profile LAR is designed to compress: high frequency, low next-token entropy, and weak coupling to task-specific parameters. We therefore test whether LAR transfers to such deployment-grade environments without modifying the agent framework itself. We instantiate this evaluation on OpenClaw, an open-source autonomous agent runtime. The LAR training pipeline is identical to that used in our main experiments (Section~\ref{sec:method}): latent actions are identified from TriviaQA rollout trajectories generated under OpenClaw, and a LoRA adapter together with new latent action embeddings is trained via trajectory-level distillation. At deployment, the learned latent tokens replace stable spans of OpenClaw's static system prompt; OpenClaw's runtime logic, tool interfaces, and ReAct loop remain unchanged. No framework-level code modification is required.

We evaluate on TriviaQA under the OpenClaw runtime and report two quantities. Compression Rate is the proportion of OpenClaw's static system-prompt tokens replaced by latent action tokens. Exact Match (EM) is the standard TriviaQA accuracy metric, measuring the fraction of agent answers matching the reference. The five settings differ only in how much of the static prompt is reparameterized: Vanilla preserves the original prompt; Short, Medium, and Long replace progressively larger contiguous spans of the static scaffolding (boilerplate, tool-format descriptions, and constraint blocks, in increasing order of semantic load); AllStatic replaces the entire static portion. The ``vs.\ Vanilla'' column reports the absolute and relative EM improvement over the uncompressed baseline.

Table~\ref{tab:openclaw_lar_results} shows that even the most conservative Short setting, compressing only 6.7\% of the static prompt, raises EM from 0.4218 to 0.5358 (a 27.0\% relative improvement). The gain is obtained purely through prompt-level reparameterization, without altering OpenClaw's tool interface, ReAct loop, or runtime, demonstrating that LAR functions as a plug-in optimization layer decoupled from the underlying agent framework. The OpenClaw Vanilla EM (0.4218) is also markedly lower than the TriviaQA accuracy in Table~\ref{tab:main_results} for the same backbone under a benchmark-native prompt, reflecting how deployment-oriented frameworks dilute task-relevant signal with structural overhead. LAR's improvement on OpenClaw is correspondingly larger, supporting the design hypothesis from Section~\ref{sec:method}: the more low-entropy structural redundancy a prompt contains, the more LAR can recover by reparameterizing it into executable latent tokens.

\begin{table}[t]
\centering
\caption{TriviaQA exact-match (EM) results of LAR applied to the OpenClaw industrial agent runtime. Compression Rate is the fraction of OpenClaw's static system-prompt tokens replaced by latent action tokens. Settings range from Vanilla (no compression) to AllStatic (full static-prompt replacement). LAR yields its largest gains under conservative compression (Short) and saturates as compression encroaches on parameter-binding content, mirroring the abstraction boundary in Section~\ref{sec:progressive}.}
\label{tab:openclaw_lar_results}
\small
\begin{tabular}{lccc}
\toprule
\textbf{Setting} & \textbf{Compression Rate} & \textbf{EM} & \textbf{vs. Vanilla} \\
\midrule
Vanilla   & 0.0\%  & 0.4218 & -- \\
Short     & 6.7\%  & \textbf{0.5358} & +0.1140 (+27.0\%) \\
Medium    & 15.2\% & 0.4672 & +0.0454 (+10.8\%) \\
Long      & 24.7\% & 0.4668 & +0.0450 (+10.7\%) \\
AllStatic & 45.3\% & 0.4308 & +0.0090 (+2.1\%) \\
\bottomrule
\end{tabular}
\end{table}

As compression deepens, gains diminish. Medium and Long retain +10.8\% and +10.7\% over vanilla, but their marginal benefit shrinks as more semantically loaded content is absorbed, and AllStatic at 45.3\% compression retains only a 2.1\% gain. This trajectory aligns with the Phase III collapse characterized in Section~\ref{sec:progressive}: once compression encroaches on segments carrying task-relevant binding information, executability is violated, and gains erode. The OpenClaw experiment therefore demonstrates that LAR can be deployed to industrial agent runtimes as a drop-in prompt-level replacement without modifying the framework itself, and also empirically reproduces the abstraction boundary predicted by our framework in a different deployment regime.


\newpage
\section*{NeurIPS Paper Checklist}


\begin{enumerate}

\item {\bf Claims}
    \item[] Question: Do the main claims made in the abstract and introduction accurately reflect the paper's contributions and scope?
    \item[] Answer: \answerYes{} 
    \item[] Justification: The abstract and introduction faithfully reflect the paper's contributions and scope. The claims about efficiency gains, preserved task performance, the abstraction boundary, and action representation as a complementary scaling axis are all supported by the methodology and experimental results presented in the paper.
    \item[] Guidelines:
    \begin{itemize}
        \item The answer \answerNA{} means that the abstract and introduction do not include the claims made in the paper.
        \item The abstract and/or introduction should clearly state the claims made, including the contributions made in the paper and important assumptions and limitations. A \answerNo{} or \answerNA{} answer to this question will not be perceived well by the reviewers. 
        \item The claims made should match theoretical and experimental results, and reflect how much the results can be expected to generalize to other settings. 
        \item It is fine to include aspirational goals as motivation as long as it is clear that these goals are not attained by the paper. 
    \end{itemize}

\item {\bf Limitations}
    \item[] Question: Does the paper discuss the limitations of the work performed by the authors?
    \item[] Answer: \answerYes{}
    \item[] Justification: The paper discusses the limitations of LAR, including its applicability conditions, failure modes, and the abstraction boundary beyond which performance degrades, in Section~\ref{sec:fail} and Section~\ref{sec:progressive}.
    \item[] Guidelines:
    \begin{itemize}
        \item The answer \answerNA{} means that the paper has no limitation while the answer \answerNo{} means that the paper has limitations, but those are not discussed in the paper. 
        \item The authors are encouraged to create a separate ``Limitations'' section in their paper.
        \item The paper should point out any strong assumptions and how robust the results are to violations of these assumptions (e.g., independence assumptions, noiseless settings, model well-specification, asymptotic approximations only holding locally). The authors should reflect on how these assumptions might be violated in practice and what the implications would be.
        \item The authors should reflect on the scope of the claims made, e.g., if the approach was only tested on a few datasets or with a few runs. In general, empirical results often depend on implicit assumptions, which should be articulated.
        \item The authors should reflect on the factors that influence the performance of the approach. For example, a facial recognition algorithm may perform poorly when image resolution is low or images are taken in low lighting. Or a speech-to-text system might not be used reliably to provide closed captions for online lectures because it fails to handle technical jargon.
        \item The authors should discuss the computational efficiency of the proposed algorithms and how they scale with dataset size.
        \item If applicable, the authors should discuss possible limitations of their approach to address problems of privacy and fairness.
        \item While the authors might fear that complete honesty about limitations might be used by reviewers as grounds for rejection, a worse outcome might be that reviewers discover limitations that aren't acknowledged in the paper. The authors should use their best judgment and recognize that individual actions in favor of transparency play an important role in developing norms that preserve the integrity of the community. Reviewers will be specifically instructed to not penalize honesty concerning limitations.
    \end{itemize}

\item {\bf Theory assumptions and proofs}
    \item[] Question: For each theoretical result, does the paper provide the full set of assumptions and a complete (and correct) proof?
    \item[] Answer: \answerNA{}
    \item[] Justification: The paper does not include formal theoretical results such as theorems or proofs. The framework introduced in the Methodology section (Section~\ref{sec:method}) defines concepts such as transition equivalence and executability as empirical constructs rather than formal theoretical claims, and they are validated through the experiments rather than proofs.
    \item[] Guidelines:
    \begin{itemize}
        \item The answer \answerNA{} means that the paper does not include theoretical results. 
        \item All the theorems, formulas, and proofs in the paper should be numbered and cross-referenced.
        \item All assumptions should be clearly stated or referenced in the statement of any theorems.
        \item The proofs can either appear in the main paper or the supplemental material, but if they appear in the supplemental material, the authors are encouraged to provide a short proof sketch to provide intuition. 
        \item Inversely, any informal proof provided in the core of the paper should be complemented by formal proofs provided in appendix or supplemental material.
        \item Theorems and Lemmas that the proof relies upon should be properly referenced. 
    \end{itemize}

    \item {\bf Experimental result reproducibility}
    \item[] Question: Does the paper fully disclose all the information needed to reproduce the main experimental results of the paper to the extent that it affects the main claims and/or conclusions of the paper (regardless of whether the code and data are provided or not)?
    \item[] Answer: \answerYes{}
    \item[] Justification: The paper provides full details necessary to reproduce the main experimental results, including the LAR algorithm, training protocol, hyperparameters, hardware setup, benchmarks, and per-task configurations in Section~\ref{sec:method}, Appendix~\ref{app:algorithm}, Appendix~\ref{app:reproducibility}, and Appendix~\ref{app:lar-config}. Anonymized source code is also released at the repository linked in the abstract.
    \item[] Guidelines:
    \begin{itemize}
        \item The answer \answerNA{} means that the paper does not include experiments.
        \item If the paper includes experiments, a \answerNo{} answer to this question will not be perceived well by the reviewers: Making the paper reproducible is important, regardless of whether the code and data are provided or not.
        \item If the contribution is a dataset and\slash or model, the authors should describe the steps taken to make their results reproducible or verifiable. 
        \item Depending on the contribution, reproducibility can be accomplished in various ways. For example, if the contribution is a novel architecture, describing the architecture fully might suffice, or if the contribution is a specific model and empirical evaluation, it may be necessary to either make it possible for others to replicate the model with the same dataset, or provide access to the model. In general. releasing code and data is often one good way to accomplish this, but reproducibility can also be provided via detailed instructions for how to replicate the results, access to a hosted model (e.g., in the case of a large language model), releasing of a model checkpoint, or other means that are appropriate to the research performed.
        \item While NeurIPS does not require releasing code, the conference does require all submissions to provide some reasonable avenue for reproducibility, which may depend on the nature of the contribution. For example
        \begin{enumerate}
            \item If the contribution is primarily a new algorithm, the paper should make it clear how to reproduce that algorithm.
            \item If the contribution is primarily a new model architecture, the paper should describe the architecture clearly and fully.
            \item If the contribution is a new model (e.g., a large language model), then there should either be a way to access this model for reproducing the results or a way to reproduce the model (e.g., with an open-source dataset or instructions for how to construct the dataset).
            \item We recognize that reproducibility may be tricky in some cases, in which case authors are welcome to describe the particular way they provide for reproducibility. In the case of closed-source models, it may be that access to the model is limited in some way (e.g., to registered users), but it should be possible for other researchers to have some path to reproducing or verifying the results.
        \end{enumerate}
    \end{itemize}

\item {\bf Open access to data and code}
    \item[] Question: Does the paper provide open access to the data and code, with sufficient instructions to faithfully reproduce the main experimental results, as described in supplemental material?
    \item[] Answer: \answerYes{}
    \item[] Justification: An anonymized version of the source code, including the latent action identification pipeline, training scripts, and evaluation harness, is publicly accessible via the repository linked in the abstract. The training protocol, hyperparameters, and per-task configurations needed to reproduce the main results are documented in Section~\ref{sec:method} and Appendix~\ref{app:reproducibility}.
    \item[] Guidelines:
    \begin{itemize}
        \item The answer \answerNA{} means that paper does not include experiments requiring code.
        \item Please see the NeurIPS code and data submission guidelines (\url{https://neurips.cc/public/guides/CodeSubmissionPolicy}) for more details.
        \item While we encourage the release of code and data, we understand that this might not be possible, so \answerNo{} is an acceptable answer. Papers cannot be rejected simply for not including code, unless this is central to the contribution (e.g., for a new open-source benchmark).
        \item The instructions should contain the exact command and environment needed to run to reproduce the results. See the NeurIPS code and data submission guidelines (\url{https://neurips.cc/public/guides/CodeSubmissionPolicy}) for more details.
        \item The authors should provide instructions on data access and preparation, including how to access the raw data, preprocessed data, intermediate data, and generated data, etc.
        \item The authors should provide scripts to reproduce all experimental results for the new proposed method and baselines. If only a subset of experiments are reproducible, they should state which ones are omitted from the script and why.
        \item At submission time, to preserve anonymity, the authors should release anonymized versions (if applicable).
        \item Providing as much information as possible in supplemental material (appended to the paper) is recommended, but including URLs to data and code is permitted.
    \end{itemize}

\item {\bf Experimental setting/details}
    \item[] Question: Does the paper specify all the training and test details (e.g., data splits, hyperparameters, how they were chosen, type of optimizer) necessary to understand the results?
    \item[] Answer: \answerYes{}
    \item[] Justification: All training and evaluation details, including data splits, optimizer, learning rate schedule, hyperparameters, hardware, and decoding settings, are specified in Section~\ref{sec:method}, Appendix~\ref{app:reproducibility}, and Appendix~\ref{app:lar-config}.
    \item[] Guidelines:
    \begin{itemize}
        \item The answer \answerNA{} means that the paper does not include experiments.
        \item The experimental setting should be presented in the core of the paper to a level of detail that is necessary to appreciate the results and make sense of them.
        \item The full details can be provided either with the code, in appendix, or as supplemental material.
    \end{itemize}

\item {\bf Experiment statistical significance}
    \item[] Question: Does the paper report error bars suitably and correctly defined or other appropriate information about the statistical significance of the experiments?
    \item[] Answer: \answerNo{}
    \item[] Justification: Due to the substantial computational cost of running multiple full-trajectory rollouts across all benchmarks, we report single-run results following common practice in LLM agent evaluation. To mitigate variance, all experiments use deterministic decoding (temperature $T=0$) and fixed random seeds, as described in Appendix~\ref{app:reproducibility}.
    \item[] Guidelines:
    \begin{itemize}
        \item The answer \answerNA{} means that the paper does not include experiments.
        \item The authors should answer \answerYes{} if the results are accompanied by error bars, confidence intervals, or statistical significance tests, at least for the experiments that support the main claims of the paper.
        \item The factors of variability that the error bars are capturing should be clearly stated (for example, train/test split, initialization, random drawing of some parameter, or overall run with given experimental conditions).
        \item The method for calculating the error bars should be explained (closed form formula, call to a library function, bootstrap, etc.)
        \item The assumptions made should be given (e.g., Normally distributed errors).
        \item It should be clear whether the error bar is the standard deviation or the standard error of the mean.
        \item It is OK to report 1-sigma error bars, but one should state it. The authors should preferably report a 2-sigma error bar than state that they have a 96\% CI, if the hypothesis of Normality of errors is not verified.
        \item For asymmetric distributions, the authors should be careful not to show in tables or figures symmetric error bars that would yield results that are out of range (e.g., negative error rates).
        \item If error bars are reported in tables or plots, the authors should explain in the text how they were calculated and reference the corresponding figures or tables in the text.
    \end{itemize}

\item {\bf Experiments compute resources}
    \item[] Question: For each experiment, does the paper provide sufficient information on the computer resources (type of compute workers, memory, time of execution) needed to reproduce the experiments?
    \item[] Answer: \answerYes{}
    \item[] Justification: The compute resources used for all experiments, including GPU type, memory, and inference setup, are reported in Appendix~\ref{app:reproducibility}.
    \item[] Guidelines:
    \begin{itemize}
        \item The answer \answerNA{} means that the paper does not include experiments.
        \item The paper should indicate the type of compute workers CPU or GPU, internal cluster, or cloud provider, including relevant memory and storage.
        \item The paper should provide the amount of compute required for each of the individual experimental runs as well as estimate the total compute. 
        \item The paper should disclose whether the full research project required more compute than the experiments reported in the paper (e.g., preliminary or failed experiments that didn't make it into the paper). 
    \end{itemize}
    
\item {\bf Code of ethics}
    \item[] Question: Does the research conducted in the paper conform, in every respect, with the NeurIPS Code of Ethics \url{https://neurips.cc/public/EthicsGuidelines}?
    \item[] Answer: \answerYes{}
    \item[] Justification: The research conforms with the NeurIPS Code of Ethics. The work involves only publicly available LLMs and benchmarks, does not involve human subjects, and raises no privacy, fairness, or dual-use concerns.
    \item[] Guidelines:
    \begin{itemize}
        \item The answer \answerNA{} means that the authors have not reviewed the NeurIPS Code of Ethics.
        \item If the authors answer \answerNo, they should explain the special circumstances that require a deviation from the Code of Ethics.
        \item The authors should make sure to preserve anonymity (e.g., if there is a special consideration due to laws or regulations in their jurisdiction).
    \end{itemize}

\item {\bf Broader impacts}
    \item[] Question: Does the paper discuss both potential positive societal impacts and negative societal impacts of the work performed?
    \item[] Answer: \answerYes{}
    \item[] Justification: As a methodological contribution to efficient LLM agent inference, LAR has potential positive impacts including reduced energy consumption and broader accessibility of agent systems through lower compute requirements. We do not identify direct negative societal impacts beyond those inherent to the underlying base LLMs, which are inherited rather than introduced by our method.
    \item[] Guidelines:
    \begin{itemize}
        \item The answer \answerNA{} means that there is no societal impact of the work performed.
        \item If the authors answer \answerNA{} or \answerNo, they should explain why their work has no societal impact or why the paper does not address societal impact.
        \item Examples of negative societal impacts include potential malicious or unintended uses (e.g., disinformation, generating fake profiles, surveillance), fairness considerations (e.g., deployment of technologies that could make decisions that unfairly impact specific groups), privacy considerations, and security considerations.
        \item The conference expects that many papers will be foundational research and not tied to particular applications, let alone deployments. However, if there is a direct path to any negative applications, the authors should point it out. For example, it is legitimate to point out that an improvement in the quality of generative models could be used to generate Deepfakes for disinformation. On the other hand, it is not needed to point out that a generic algorithm for optimizing neural networks could enable people to train models that generate Deepfakes faster.
        \item The authors should consider possible harms that could arise when the technology is being used as intended and functioning correctly, harms that could arise when the technology is being used as intended but gives incorrect results, and harms following from (intentional or unintentional) misuse of the technology.
        \item If there are negative societal impacts, the authors could also discuss possible mitigation strategies (e.g., gated release of models, providing defenses in addition to attacks, mechanisms for monitoring misuse, mechanisms to monitor how a system learns from feedback over time, improving the efficiency and accessibility of ML).
    \end{itemize}
    
\item {\bf Safeguards}
    \item[] Question: Does the paper describe safeguards that have been put in place for responsible release of data or models that have a high risk for misuse (e.g., pre-trained language models, image generators, or scraped datasets)?
    \item[] Answer: \answerNA{}
    \item[] Justification: The paper does not release any new pretrained models, datasets, or generative systems that pose a high risk of misuse. LAR is a parameter-efficient adaptation method (LoRA adapter and latent action embeddings) applied to existing publicly released base models.
    \item[] Guidelines:
    \begin{itemize}
        \item The answer \answerNA{} means that the paper poses no such risks.
        \item Released models that have a high risk for misuse or dual-use should be released with necessary safeguards to allow for controlled use of the model, for example by requiring that users adhere to usage guidelines or restrictions to access the model or implementing safety filters. 
        \item Datasets that have been scraped from the Internet could pose safety risks. The authors should describe how they avoided releasing unsafe images.
        \item We recognize that providing effective safeguards is challenging, and many papers do not require this, but we encourage authors to take this into account and make a best faith effort.
    \end{itemize}

\item {\bf Licenses for existing assets}
    \item[] Question: Are the creators or original owners of assets (e.g., code, data, models), used in the paper, properly credited and are the license and terms of use explicitly mentioned and properly respected?
    \item[] Answer: \answerYes{}
    \item[] Justification: All datasets (TriviaQA, HumanEval, MBPP, KodCode, Mind2Web) and base models (Meta-Llama-3.1-8B-Instruct, Qwen3-8B) used in this paper are publicly released for research use, and their original papers are cited at first mention. We use these assets in accordance with their respective licenses and terms of use.
    \item[] Guidelines:
    \begin{itemize}
        \item The answer \answerNA{} means that the paper does not use existing assets.
        \item The authors should cite the original paper that produced the code package or dataset.
        \item The authors should state which version of the asset is used and, if possible, include a URL.
        \item The name of the license (e.g., CC-BY 4.0) should be included for each asset.
        \item For scraped data from a particular source (e.g., website), the copyright and terms of service of that source should be provided.
        \item If assets are released, the license, copyright information, and terms of use in the package should be provided. For popular datasets, \url{paperswithcode.com/datasets} has curated licenses for some datasets. Their licensing guide can help determine the license of a dataset.
        \item For existing datasets that are re-packaged, both the original license and the license of the derived asset (if it has changed) should be provided.
        \item If this information is not available online, the authors are encouraged to reach out to the asset's creators.
    \end{itemize}

\item {\bf New assets}
    \item[] Question: Are new assets introduced in the paper well documented and is the documentation provided alongside the assets?
    \item[] Answer: \answerYes{}
    \item[] Justification: The new assets introduced in this paper, including the latent action identification pipeline, training scripts, evaluation harness, and trained LoRA adapters with latent action embeddings, are released in the anonymized repository linked in the abstract, with documentation describing usage, configuration, and reproduction of the main results.
    \item[] Guidelines:
    \begin{itemize}
        \item The answer \answerNA{} means that the paper does not release new assets.
        \item Researchers should communicate the details of the dataset\slash code\slash model as part of their submissions via structured templates. This includes details about training, license, limitations, etc. 
        \item The paper should discuss whether and how consent was obtained from people whose asset is used.
        \item At submission time, remember to anonymize your assets (if applicable). You can either create an anonymized URL or include an anonymized zip file.
    \end{itemize}

\item {\bf Crowdsourcing and research with human subjects}
    \item[] Question: For crowdsourcing experiments and research with human subjects, does the paper include the full text of instructions given to participants and screenshots, if applicable, as well as details about compensation (if any)? 
    \item[] Answer: \answerNA{}
    \item[] Justification: This work does not involve crowdsourcing or research with human subjects.
    \item[] Guidelines:
    \begin{itemize}
        \item The answer \answerNA{} means that the paper does not involve crowdsourcing nor research with human subjects.
        \item Including this information in the supplemental material is fine, but if the main contribution of the paper involves human subjects, then as much detail as possible should be included in the main paper. 
        \item According to the NeurIPS Code of Ethics, workers involved in data collection, curation, or other labor should be paid at least the minimum wage in the country of the data collector. 
    \end{itemize}

\item {\bf Institutional review board (IRB) approvals or equivalent for research with human subjects}
    \item[] Question: Does the paper describe potential risks incurred by study participants, whether such risks were disclosed to the subjects, and whether Institutional Review Board (IRB) approvals (or an equivalent approval/review based on the requirements of your country or institution) were obtained?
    \item[] Answer: \answerNA{}
    \item[] Justification: This work does not involve human subjects, so IRB approval is not required.
    \item[] Guidelines:
    \begin{itemize}
        \item The answer \answerNA{} means that the paper does not involve crowdsourcing nor research with human subjects.
        \item Depending on the country in which research is conducted, IRB approval (or equivalent) may be required for any human subjects research. If you obtained IRB approval, you should clearly state this in the paper. 
        \item We recognize that the procedures for this may vary significantly between institutions and locations, and we expect authors to adhere to the NeurIPS Code of Ethics and the guidelines for their institution. 
        \item For initial submissions, do not include any information that would break anonymity (if applicable), such as the institution conducting the review.
    \end{itemize}

\item {\bf Declaration of LLM usage}
    \item[] Question: Does the paper describe the usage of LLMs if it is an important, original, or non-standard component of the core methods in this research? Note that if the LLM is used only for writing, editing, or formatting purposes and does \emph{not} impact the core methodology, scientific rigor, or originality of the research, declaration is not required.
    \item[] Answer: \answerYes{}
    \item[] Justification: LLMs are central to this work as the agent backbones being studied. Specifically, Meta-Llama-3.1-8B-Instruct and Qwen3-8B serve as the base models on which LAR is trained and evaluated, as described in Section~\ref{sec:method} and Appendix~\ref{app:agent-models}. No LLM was used as an original or non-standard component beyond serving as the experimental subject. LLMs were also used to assist with writing and grammar polishing, which does not impact the core methodology or scientific rigor of this work.
    \item[] Guidelines:
    \begin{itemize}
        \item The answer \answerNA{} means that the core method development in this research does not involve LLMs as any important, original, or non-standard components.
        \item Please refer to our LLM policy in the NeurIPS handbook for what should or should not be described.
    \end{itemize}

\end{enumerate}

\end{document}